\pdfoutput=1
\documentclass{ceurart}

\usepackage[utf8]{inputenc}
\usepackage[english]{babel}
\usepackage{graphicx, caption}
\usepackage{amsmath}
\usepackage{amssymb}
\usepackage{booktabs}
\usepackage{longtable}
\usepackage{array}
\usepackage[onehalfspacing]{setspace}

\graphicspath{{figures/}}
\newcommand{\productionfigurewidth}{0.98\textwidth}
\newcommand{\methodschemascale}{0.72}

\RenewDocumentCommand \logonote { m } {}

\begin{document}

\conference{Preprint. Submitted to \textit{Frontiers in Complex Systems}.}

\title{Tracing individual knowledge trajectories in a changing field: the case of general relativity and gravitation}

\author[1]{Raphael Schlattmann}[%
email=raphael.schlattmann@tu-berlin.de,
orcid=0000-0002-2234-3720
]
\cormark[1]

\author[2]{Malte Vogl}[%
email=malte.vogl@bbaw.de,
orcid=0000-0002-2683-6610
]

\address[1]{Technische Universit\"at Berlin, Berlin, Germany}
\address[2]{Berlin-Brandenburgische Akademie der Wissenschaften, Berlin, Germany}

\cortext[1]{Corresponding author.}

\begin{abstract}
Historians have reconstructed the twentieth-century transformation of general relativity and gravitation (GRG) both at the field level and through individual careers, but connecting these scales requires a way to compare researchers with the changing field over time. We develop such a comparison for individual researchers or groups. Across successive two-year periods, their publications and references are set against GRG field literature from the same, earlier, and later periods. Building on an earlier study's definition of \textit{Own Vocabulary}, an information-theoretic comparison of author and field language, and \textit{Embedding Density Estimation}, which locates an author's semantic neighbourhood in document-embedding space~\citep{schlattmann_trajectories_2024}, we extend the analysis to the fifty most-published authors in a NASA/ADS corpus of roughly 180,000 GRG records (1911 to 2000) and add two citation-based measures: \textit{Referenced Vocabulary}, following the language of cited works, and \textit{Citation Identity}, following which authors are co-cited. Together, the four measures compare an author's written language, cited literature, semantic neighbourhood, and configuration of cited authorities with the surrounding field. The earlier cases of Joseph Silk and Hans-Jürgen Treder suggested that closer alignment with field vocabulary accompanies a denser semantic neighbourhood. We find that across the most-published authors this relationship holds only partially. Written and cited vocabularies tend to move together, usually resembling later GRG literature as the field turned increasingly towards astrophysical and cosmological research. Semantic neighbourhoods, by contrast, more often lie where the field's publications were concentrated in earlier periods, while co-citation patterns follow no single temporal direction and the two citation measures frequently place the same researcher differently, despite drawing on identical reference lists. Individual trajectories can thus combine vocabulary tied to later field states with older semantic or citation structures, and these divergent trajectories classify certain patterns for closer historical investigation. The approach is transferable to other fields, for which defensible corpus boundaries and adequate coverage of texts, references, as well as disambiguated author identities can be provided.
\end{abstract}


\maketitle

\section{Introduction}

For research on general relativity and gravitation (GRG) in the twentieth century, historians have reconstructed how dispersed agendas became a connected research field that was organised first around general relativity proper and then shifted more towards relativistic astrophysics \citep{eisenstaedt_low_1989,blum_reinvention_2015,lalli_dynamics_2020}. This phase of field formation, the so-called Renaissance of General Relativity, is, from a quantitative perspective, a macrohistorical or structural phenomenon which must be reconciled systematically with individual or microhistories if their integration is to be taken seriously \citep{renn_evolution_2020}. This article connects these two spheres by comparing the publications of individual researchers with those of the changing field period by period (synchronous and asynchronous). It asks how close or distant field-level transformation appears in an individual publication's language, cited literature, and semantic location, and whether these relations change together or when and where they depart.

At the base of this method lies the approach of socio-epistemic networks (SEN), which treats systems of knowledge, such as scientific fields, as interconnected and changing relations between researchers (social), cognitive (semantic) and material (semiotic) knowledge representations \citep{renn_netzwerke_2016,renn_evolution_2020,kaye_socio-epistemic_2024}. Here we focus mainly on the latter two. An individual's relation to the field may look different in language, references, or semantic position. A researcher may use language characteristic of an emerging field while citing older literature or may be working on topics formed earlier. We reconstruct these relations from publications, the material layer in SEN terms. In each two-year slice of a given time window, we compare one feature of an author's publications with the same feature in the rest of the GRG corpus from the same, earlier, or later slices. Reading every author against this moving field baseline renders the authors comparable with respect to their (changing) position in a changing field.

Our earlier GRG study put part of this comparison into practice for Joseph Silk and Hans-Jürgen Treder \citep{schlattmann_trajectories_2024}. The measure Own Vocabulary used relative entropy to compare each author's language with the field's. Embedding Density Estimation (EDE) measured how densely field publications occupied the semantic neighbourhood around that author's work. We found that Silk's growing alignment with the field's growing astrophysical and cosmological vocabulary was accompanied by a densification around his publications. On the other hand Treder's growing lexical distance developed parallel to a decreasing density around his more principle-oriented publications. The contrast suggested that a development towards closer field-language alignment may accompany a development towards greater semantic density, e.g., topic alignment, and vice versa. Two cases could not establish whether this relationship recurred systematically across the field, and the relation with citation practice remained untested.

This article evaluates this hypothesis systematically and adds two citation measures. Referenced Vocabulary asks whether the language of an author's cited works resembles the literature cited by the field in the same, earlier, or later periods. Citation Identity makes the corresponding comparison for pairs of authors cited together \citep{small_co-citation_1973,white_authors_2001}. Although both measures begin with the same reference lists, one follows the language of cited literature and the other the configuration of cited authorities. We compare them with Own Vocabulary and Embedding Density Estimation for the fifty authors (disambiguated identities) with the most publications in a NASA/ADS corpus of about 180,000 GRG records from 1911 to 2000.

The results provide only partial support for the hypothesis. We find that written and cited vocabularies move together most clearly and usually resemble later field states. Semantic neighbourhoods follow a different path and more often lie where the field's publications were concentrated in earlier periods. The new citation measures show that works written in the language of a later field state can be cited without reproducing the configuration of authorities found in that field. Individual trajectories can therefore combine later-field language with an older semantic neighbourhood or a distinct citation structure. We give historical weight to patterns that recur across representations and survive the robustness checks. Authors whose measures remain apart become cases for closer historical study.

\begin{table}[ht!]
\caption{The four measures, their target (the publications of the individual(s)) and field (the publications of the rest of the corpus) traces, and the relation represented.}
\label{tab:argument-overview}
\begin{center}
\footnotesize
\begin{tabular}{@{}>{\raggedright\arraybackslash}p{0.18\textwidth}>{\raggedright\arraybackslash}p{0.22\textwidth}>{\raggedright\arraybackslash}p{0.22\textwidth}>{\raggedright\arraybackslash}p{0.26\textwidth}@{}}
\hline
Measure & Target Feature& Field Feature& Relation\\
\hline
Own Vocabulary & Terms in the target's own titles + abstracts& Terms in field titles + abstracts& Written vocabulary relative to the field\\
Referenced Vocabulary & Terms in the target's cited works' titles + abstracts& Terms in the field's cited works' titles + abstracts& Cited vocabulary relative to the field\\
Citation Identity & Co-cited author pairs in the target's reference lists & Co-cited author pairs in field's reference lists& Configuration of authorities relative to the field\\
Embedding Density Estimation & Document embeddings of target publications& Document embeddings of field publications& Field occupancy of the semantic neighbourhood around the targets publications\\
\hline
\end{tabular}
\end{center}
\end{table}

\section{Related Work}

\subsection{The renaissance of general relativity in current historiography}

The twentieth-century history of general relativity is often described as a recovery. After the confirmation of light bending in 1919, the theory supposedly entered a long period that Jean Eisenstaedt called its ``low-water mark'', running from the mid-1920s to the mid-1950s, when general relativity was treated as mathematically demanding, physically marginal and was studied by few \citep{eisenstaedt_low_1989}. The years that followed are usually called a renaissance \citep{will_renaissance_1989,blum_renaissance_2016}, and Kip Thorne placed a ``golden age'' of the theory at roughly 1960 to 1975, when it ``entered the mainstream of theoretical physics'' \citep[p.~74]{thorne_black_1994}.

Whether and how this recovery occurred is itself a historiographical question, and \citet[pp.~3--4]{blum_renaissance_2020} survey competing answers. An observation-first account credits the astronomical discoveries of the 1960s, which demonstrated that the theory ``would have important applications in astrophysical situations'' \citep[p.~16]{will_renaissance_1989}. Yet Peebles dates the renaissance of experimental gravity to Robert Dicke's programme, announced at Chapel Hill in 1957, before quasars or the microwave background were known \citep[p.~3]{peebles_dicke_2017}. A patronage account prioritises funding and influx of people rather than discoveries. Private patrons, Roger Babson and Agnew Bahnson chief among them, financed American gravity research while the field was still marginal, military funding followed, and the Cold War expansion of theoretical physics supplied further personnel \citep{kaiser_price_2018,blum_renaissance_2020}. A third family of accounts points towards theoretical developments. As one example \citet[p.~614]{blum_reinvention_2015} point at the Petrov classification of 1954, developed at the margins in Kazan, which was taken up by Felix Pirani for gravitational waves and became an essential tool on the way to the Kerr solution. Also some established physicists from other fields began turning toward gravitation, e.g. John Wheeler, whose move Blum and Brill read as the recognition of the theory's ``untapped potential'' \citep[p.~142]{blum_tokyo_2020}.

More recent revisions, or rather reframing, of those three accounts starts at the low-water mark period itself. Rather than a period of absolute stagnation, it is now mostly understood as an era of social and epistemic dispersion \citep{lalli_building_2017,blum_reinvention_2015,blum_gravitational_2018,lalli_dynamics_2020,lalli_socio-epistemic_2020}. Research continued on relativistic cosmology, unified field theories and the quantisation of the field equations, but it lacked connection. \citet[p.~5]{blum_renaissance_2020} describe the period as one of ``strong dispersion of research agendas related to Einstein's theory, all of which were becoming marginalized with respect to major advances in physics'' \citep{kragh_cosmology_1996,kennefick_traveling_2007}. The same theory served cosmologists, quantum physicists, mathematicians, and the few astronomers who tried to test it, but their respective uses rarely intersected before the post-war community-building efforts bridged these divides.

But the connection was built before the rise of relativistic astrophysics, which has often been given the credit. \citet[p.~601]{blum_reinvention_2015} call this a ``reinvention of general relativity, which was turned by these dynamics from a theoretical framework into a field of study in its own right''. On this reading the above narratives describe one process from different angles. The dispersed agendas of the low-water mark were drawn together around problems internal to the theory, gravitational waves and the nature of singularities among them, and the community forming around those problems built the shared tools and standards that the separate uses had lacked. Andr\'e Mercier's 1955 Bern jubilee conference was, in \citet{lalli_building_2017}'s reading, an explicit effort to build a community where none yet existed. Chapel Hill in 1957 gave a younger, empirically minded generation a forum, and the 1959 Royaumont meeting led to the International Committee on GRG and its conference series. What turned meetings into an integration was circulation, young researchers moving between centres as students or postdocs and transferring problems and methods across groups that the conferences had only brought into contact \citep[p.~26]{lalli_dynamics_2020}. 

The socio-epistemic networks (SEN) approach follows just this kind of formation across social, material, and semantic network layers \citep{lalli_socio-epistemic_2020,renn_evolution_2020}. For the social layer, \citet[p.~1160]{lalli_dynamics_2020} show that exactly this mobility and community building produced a giant component in the collaboration network between 1959 and 1960, ``well after the end of World War II, but before the discovery of quasars''. In the semantic and material layers, co-citation moved away from programmes that sought to replace or extend Einstein's theory and toward a core concerned with general relativity proper \citep{lalli_socio-epistemic_2020}. The astrophysical discoveries of the 1960s then added another topical centre to a field already taking shape \citep{blum_gravitational_2018}.

This network-based addition has sharpend what historiography should explain, for example against objections like Goenner's, who argues that research grew rather steadily after 1945, that the earlier stagnation reflects a lack of funding and manpower that generally applied to most of physics, in other words that no "renaissance'' took place, or that the label at least rests on weak empirical grounding \citep{goenner_general_2016,goenner_golden_2017}. But mere growth in numbers cannot explain why previously separate agendas connected when and how they did. The network shift of 1959-1960 and the co-citation turn toward general relativity proper do not prove a renaissance in every sense of that label. But they specify the object that changed, which is the active building of structural connectivity among continuing problem traditions and the standards of relevance binding them. It is in this sense that GRG became a field.

The community that formed also drew the boundary of its own field. From 1962 the \textit{Bulletin on General Relativity and Gravitation}, edited by Mercier, listed active researchers and the topics they claimed, an undertaking that expressed, in Lalli's reading, ``the conscious intent to build a community'' in a field ``at the time both dispersed and undefined'' \citep[p.~57]{lalli_building_2017}. The \textit{Bulletin} thus provides evidence that the formation of GRG involved not only growing connectivity, but also an explicit delineation of the field.

The combination of frame-based historiography and historical network analysis share an ambition, which Renn understands as linking micro- and macro-history in the study of knowledge \citep[p.~319]{renn_evolution_2020}. \citet[pp.~16--17]{lalli_socio-epistemic_2020} name the connection between the social, semiotic, and semantic network layers as the goal of their programme while confining their analysis, for the GRG case, to the structure within each layer. What is still missing between the two scales are quantitative instruments that state, in comparable units, how an individual's trajectory developed relative to the change that the structural or field-level studies describe. Our earlier study built a first pair of such instruments for two authors \citep{schlattmann_trajectories_2024}. What the broader quantitative traditions offer for this relational task and where they stop short of it is the subject of the next section.

\subsection{Quantitative approaches to individual scientific trajectories}

Measuring a trajectory against the development of a field requires operationalizing both sides of the comparison together, the individual and the field. But quantitative approaches that look at similar problem spaces mostly retain only one of them.

A first strand, the science of science, usually studies careers in large bibliometric datasets and treats an individual trajectory as a sequence of an author's publications, asking what statistical regularities govern its shape \citep{fortunato_science_2018,wang_science_2021}. For example \citet{sinatra_quantifying_2016} showed that the highest-impact paper position of an author is uniform in this sequence, so the most influential work of a career path may come early, midway or late. \citet{liu_hot_2018} refined this picture by observing that an author's several most cited works nevertheless tend to fall within a few years of one another, a clustering they termed \textit{hot streaks}. For our purposes, the decisive feature of these results is their frame of reference. Each of those regularities is measured against the career trajectory itself, as a position or a pattern within an author's own sequence of works, but the generalizability of these findings, which extend to most careers \citep{li_elite_2020}, rests on removing what is specific to a field. Impact and productivity are normalised across fields and cohorts so that careers become comparable in the first place. The field therefore enters the analysis as a background to be subtracted rather than as an object with a state of its own. Whether a given physicist moved relative to e.g. the astrophysical turn of GRG is not a quantitative question that these designs pose. The question concerns the state of one field at one time, and that state is what normalisation removes. 

A second strand of quantitative work maps fields rather than careers. Author co-citation analysis, introduced by White and McCain and since developed into a standard instrument of science mapping, counts how often two authors' works are cited together and creates a discipline's intellectual map from those pairings \citep{white_visualizing_1998,chen_science_2017,voglTemporalitiesSpreadKnowledge2025b}. An author's position on such a map is a location in the aggregate reception of literature, fixed by how the citing community as a whole combines authorities. \citet{white_authors_2001} later proposed citer analysis as a complementary perspective, in which a single oeuvre is read through the references its author repeatedly makes. This is considerably closer to our objective, since the unit of description is now one person's citing behaviour rather than the community's reception. A related aggregate perspective on change over time comes from computational studies of lexical semantic change, which infer the drift of a term's meaning from the shift of its neighbourhood in embedding spaces trained on successive periods of a corpus \citep{hamilton_diachronic_2016,schlechtweg_semeval_2020,tahmasebi_computational_2026}. Later work in this tradition replaced static period-wise vectors with contextualised representations of word occurrences, which sharpened the measurement without changing its object \citep{giulianelli_analysing_2020,kutuzov_contextualized_2022,periti_systematic_2024}. The unit of change remains the word in the language at large. What all of these approaches deliver are positions in aggregate structures, whether topography drawn from everyone's citations or a drift measured in language. But even citer analysis, which does isolate the individual, describes the oeuvre in its own terms and not relative to the state of a surrounding field. None of these approaches expresses a person's trace as a deviation/alignment from the reconstructed state of a specific field in a specific period.

The instruments of this article descend from a third strand, which compares corpora instead of mapping them. In a series of studies of the Royal Society Corpus, \citet{degaetano-ortlieb_using_2018,bizzoni_linguistic_2020} used relative entropy between time slices of corpora to detect periods of linguistic change without fixing the periods in advance, to separate typical from productive usage \citep{degaetano-ortlieb_information-based_2016}, and to describe the long-run conventionalisation of scientific English \citep{teich_less_2021,degaetano-ortlieb_toward_2022}. Two probability distributions are estimated over the same features, one for each body of text, and an asymmetric divergence measure states how far the first stands from the second, so that no period boundaries and no external reference points have to be assumed. The comparison is therefore between two reconstructed states, although in this tradition both states are aggregates, sections of the same shared corpus rather than subsets like individuals. Our earlier study turned this corpus-against-corpus idea into an author-against-field design for two physicists, estimating one distribution from an author's publications and the other from the field corpus of the same period and extending it to asynchronous comparison \citep{schlattmann_trajectories_2024}. This article extends our earlier design from two measures to four, and from two authors to fifty.

\section{Materials and Methods}

\subsection{The NASA/ADS GRG corpus}

The corpus under investigation is a set of publications related to GRG drawn from the \href{https://ui.adsabs.harvard.edu/}{NASA Astrophysics Data System} (ADS), covering 1911 to 2000. We follow the selection logic of the historical work on the renaissance \citep{blum_renaissance_2020,lalli_dynamics_2020} but extend the period and use ADS, which covers GRG more broadly than the Web-of-Science samples of earlier network studies and includes more non-English work. The ADS query strategy follows the set-based construction of the previous study \citep{schlattmann_trajectories_2024}. See Supplementary Material for preparation and cleaning rules, translation handling, per-author coverage, and the embedding, projection, and aggregation settings, alongside the public code and configuration files. 

The final corpus holds about 180,000 publications. Author identities were disambiguated with a purpose-built pipeline. The \href{https://github.com/raphschlatt/ads-bib}{\texttt{ads-bib}} python package turns ADS records into curated datasets, and the \href{https://github.com/raphschlatt/ads-and}{\texttt{ads-and}} package resolves author-name variants into stable identifiers (see the Data Availability Statement). The disambiguated identifiers are the analytical identity layer throughout this paper. The vocabulary and density measures use publication titles concatenated with abstracts, machine-translated into English where needed. The citation measures use reference metadata where they are available. 

This corpus is a historically formed publication space, not a complete record of GRG research. Its coverage is uneven across languages, journals, periods, and document types, and reference metadata is sparser for non-English and less indexed journals. Later years hold more publications, partly through real growth and partly through collection effects. These asymmetries also shape which authors are visible and how far their trajectories can be read. Treder is a strong example of these collection effects. References in his publications, for example, appear in the corpus at roughly a tenth of the rate found in a manual reconstruction of his complete output, while most of his publications are indexed \citep{schlattmann_trajectories_2024}. 

\subsection{Targets, field baselines, and time slices}

The basic unit of comparison relates a target to a field within a time slice. The target is the set of publications by an author, group, institution, or other historically meaningful subset. The field is the GRG publication space in the same period, but with the target removed when a measure requires it. Time slices are chosen to be two years wide and do not overlap, so each target can be compared with both the contemporary field (synchronous) and field states from earlier or later periods (asynchronous). Time slices are practical historical constructions and locate publication traces, not necessarily moments in which ideas were formed.

The systematic analysis is author-centred, selecting authors by their disambiguated identifiers rather than name matching. Authors are ranked by the number of corpus publications carried by one disambiguated identity, after placeholder identities are removed, and the Top-50 are the fifty identities with the most corpus publications. Selecting by coverage also secures the temporal spread a trajectory needs and more publications tend to correlate with longer trajectories (the Top-50 range from nine to twenty-six active two-year slices, see Supplementary Material). The resulting set is one of corpus internal visibility in which trajectories of very different shape can be compared with the same measures, with Silk and Treder kept as reference points from the earlier study. Selection by corpus coverage measures visibility in the curated publication space, not necessarily historical visibility, and the two need not coincide but do in most of our cases. The portfolio therefore includes a variety of profiles from highly cited cosmologists to experimental specialists, and more locally received researchers.

Where an author's closest field state falls is an empirical result rather than a built-in window. In practice the vocabulary measures match field states within roughly a decade of an author's own slices, while the reference-based measures and especially Embedding Density can reach back to much earlier field states. A large negative lead/lag (defined in Section~\ref{sec:divergence}) therefore means resemblance to a much earlier field state, not a location outside the period in which these authors published. Detailed preparation rules, run settings, and reproducibility information are documented in the Supplementary Material.

\subsection{Measuring an author's relation to the field}

We compare an individual's trajectory against the field along four modes. The first one is language used by the authors themselves (Own Vocabulary). As GRG moved toward general relativity proper and then more toward relativistic astrophysics, a vocabulary of unification and principle gave way to one of radiation, collapse, singularities, and observational cosmology \citep{blum_reinvention_2015,kragh_cosmology_1996}. The second is the language used in cited literature (Referenced Vocabulary), since an author can write using the field's contemporary vocabulary while drawing on older textual resources that might use a different language. The third is semantic density (Embedding Density Estimation), because some combinations of problems or topics became at certain points in time crowded with publications, while others thinned. And the fourth is co-citation practice (Citation Identity), since authors and field also steadily rearranged which earlier works were cited together \citep{lalli_socio-epistemic_2020,small_co-citation_1973}.

The measures that follow begin from standard publication traces of papers with title, abstract, and reference list. Titles and abstracts supply terms, while reference lists point to cited works, their titles and abstracts but also to the authors who appear together in these lists. Each measure turns one of these traces into a profile for a target and a matching profile for the surrounding field and asks how far the two stand apart. Because every target is read against the same field profile, authors are compared on the same scale. Throughout this section we write \(T\) for the target and \(F\) for the field. Each measure comes with specific design choices, e.g., to prevent over-inflation by papers with many references (see the design-choice table in the Supplementary Material). For our interpretation of what synchronous and asynchronous results from different measures refer to, see Section~\ref{sec:relationmeasures}.

\subsection{The target--field divergence}
\label{sec:divergence}

How far does an author's usage of some feature stand from what the field of a given moment would lead one to expect\footnote{We call this reference point the ``mainstream'' of a slice, as in the earlier study \citep{schlattmann_trajectories_2024}. The term is operational. It names the pooled features of all other publications in the slice, in which frequent features dominate, and it carries no claim about the standing or reception of those who use them. Whether such a pool also marks a mainstream in the historical sense is a question the results address, and the measures do not presuppose it.}? Three of the four measures we introduce here are based on the same information theoretic approach. We represent both a target and the surrounding field as probability distributions over a shared set of features \(V\), where a feature is a term in the two vocabulary measures (\textit{Own and Referenced Vocabulary}) or a pair of co-cited authors in Citation Identity, and each feature carries a probability from its relative frequency in the slice. Following the approach of the earlier study \citep{schlattmann_trajectories_2024}, we build one distribution for the target in a two-year slice \(s\) and one for the field in a two year slice \(\tau\), and we measure how far the target diverges from the field via Kullback--Leibler divergence (KLD),
\[
D_{\mathrm{KL}}\big(P_T(\cdot \mid s) \parallel P_F(\cdot \mid \tau)\big)
= \sum_{x \in V} P_T(x \mid s)\,\log_2 \frac{P_T(x \mid s)}{P_F(x \mid \tau)} ,
\]
where \(P_T(x \mid s)\) and \(P_F(x \mid \tau)\) are the probabilities of feature \(x\) in the target and field. KLD is the average number of additional bits needed to encode the target's features using the field's distribution, so a lower value means the two are more alike. The comparison is deliberately asymmetric. It measures the target against the field's expectations, so a feature weighs in proportion to how often the target uses it. Because KLD is a sum, the divergence also records which features drive it most. A field-wide term such as ``relativity'' barely distinguishes anyone, while a term tied to a single programme can mark a clear difference \citep{degaetano-ortlieb_information-based_2016}; see Supplementary Material for an example.

A distribution assigns zero probability to any feature it has not seen, and a single zero makes the divergence infinite. In a historically uneven corpus this is common\footnote{For terms this is called the out-of-vocabulary (OOV) problem}. Early slices for example have less publications overall but are also less-indexed, which might leave many features unobserved. But absence from weakly indexed slices is not the same as historical impossibility in the field. We therefore smooth each distribution with Jelinek--Mercer smoothing \citep{jelinek_interpolated_1980,zhai_study_2004}, mixing the slice distribution with a background distribution \(P_C\) estimated over the whole field,
\[
\widetilde P_T(x \mid s) = (1-\lambda)\,P_T(x \mid s) + \lambda\,P_C(x) ,
\]
and likewise for the field. Every feature then keeps a small non-zero probability, and a single slice is read against the background of the field as a whole. We set the interpolation weight to \(\lambda = 0.05\) in the main analysis, following the earlier study \citep{schlattmann_trajectories_2024}, so that each slice is read mostly from its own distribution, and we apply a probability floor \(\varepsilon = 10^{-12}\) before renormalising. The setting \(\lambda = 0.5\), which mixes slice and field background evenly, serves as a robustness axis whose effect we report among the robustness boundaries below. The divergence above is computed on the smoothed distributions \(\widetilde P_T\) and \(\widetilde P_F\).

We use the same divergence in two ways. A synchronous comparison sets \(\tau = s\) and measures how far a target stands from the field of its own moment. An asynchronous comparison weighs one target slice against the field of every available period and records the closest one, reported as a lead/lag \(\Delta = \tau^\ast - s\). If for example a target's 1970 slice is closest to the field slice of 1958, the negative lead/lag points to an earlier field state. See Section~\ref{sec:warrant} on the historical interpretation of these differences.

For the three distributional measures, a pooled time slice of x publications can hide whether a divergence is attributable to only one, a few or all of those x publications. A feature may drive the divergence because it runs through most of a target's publications or because it saturates a single long or unusual one, and usually these cases matter historically. For example a term spread across many papers is closer to a durable research interest, while a term concentrated in one publication may mark a single project. We therefore test the strongest divergence drivers with a document-level Welch t-test \citep{welch_generalization_1947}, as a stability check on the all-term result. We correct these tests for multiple comparisons with the Benjamini--Hochberg procedure \citep{benjamini_controlling_1995}, holding synchronous tests within a \texttt{slice} and asynchronous tests within a \texttt{target\_slice}. Per-document support is sparse for many authors and slices, and a strict false-discovery threshold would retain almost no features. We therefore read this layer as an exploratory set of candidate features, retained at a false-discovery rate of \(q \le 0.20\). The all-term lead/lag results are computed on the full distributions without threshold.

\subsubsection{Own Vocabulary}
\label{sec:ownvocabulary}

\begin{figure}[ht!]
\begin{center}
\includegraphics[scale=\methodschemascale]{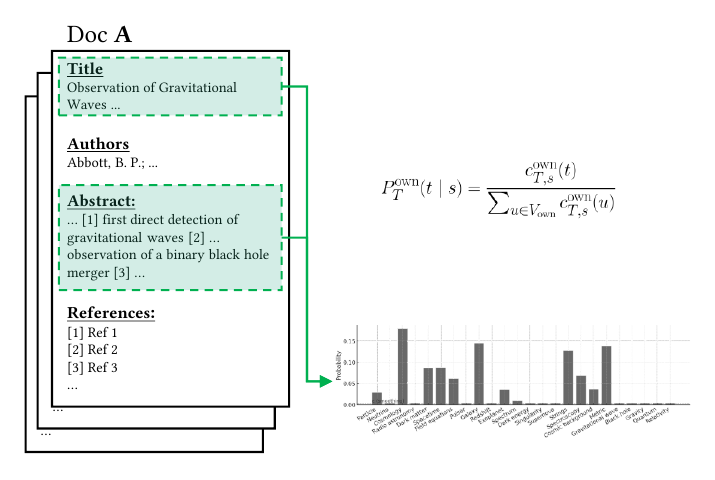}
\end{center}
\caption{Own Vocabulary, the first step of a single worked example carried through Figures~\ref{fig:method-ref}--\ref{fig:method-ede}. The lemmatised terms in a target publication's own title and abstract form its term profile \(P_T^{\mathrm{own}}(t \mid s)\), shown as a distribution over terms; the field profile is built the same way from the surrounding publications. The document shown is a stylised illustration of the construction, not a corpus record.}
\label{fig:method-own}
\end{figure}

\textit{Own Vocabulary} asks whether an author uses the same vocabulary as the contemporary field or that of an earlier or later field state. It builds the target distribution from the terms in the author's own titles and abstracts. For a target \(T\) in slice \(s\) we count how often each lemmatised term \(t\) occurs across the target's publications, written \(c^{\mathrm{own}}_{T,s}(t)\), and set its probability to the term's share in that slice,
\[
P_T^{\mathrm{own}}(t \mid s)
= \frac{c^{\mathrm{own}}_{T,s}(t)}{\sum_{u \in V_{\mathrm{own}}} c^{\mathrm{own}}_{T,s}(u)} ,
\]
with \(V_{\mathrm{own}}\) the shared set of terms. The field distribution \(P_F^{\mathrm{own}}(t \mid \tau)\) is built the same way from the field's publications, and the two are compared via KLD as shown above. Figure~\ref{fig:method-own} shows the construction for a single target publication, whose title and abstract supply the terms that enter the profile (``gravitational'', ``wave'', ``radiation''). 

This is the most direct trace of an author's work, the language in which they pose their own problems, and it is historically informative because the field's language evolves. A paper written around ``collapse'', ``singularity'', and ``horizon'' uses a different vocabulary from one written around ``unified'', ``field'', and ``Mach''. A low divergence marks an author writing in the current field idiom, a high one a distinctive vocabulary.

\subsubsection{Referenced Vocabulary}

\begin{figure}[ht!]
\begin{center}
\includegraphics[scale=\methodschemascale]{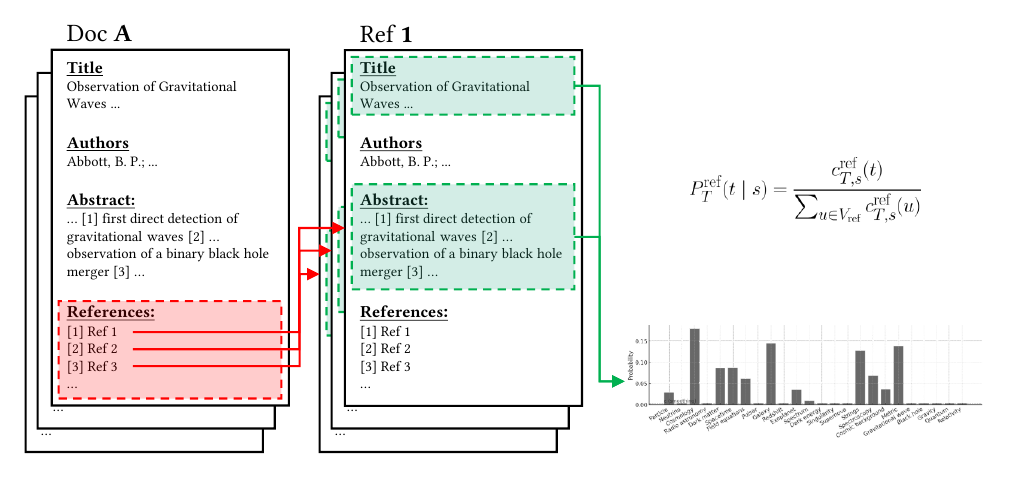}
\end{center}
\caption{Referenced Vocabulary, the same worked example at the next layer. The titles and abstracts of the works the same target publication cites supply the terms of its referenced-vocabulary profile \(P_T^{\mathrm{ref}}(t \mid s)\), weighted across citing documents and cited works.}
\label{fig:method-ref}
\end{figure}

\textit{Referenced Vocabulary} asks which vocabulary the literature an author cites curates, measured against the vocabulary the field curates via citations in the same, earlier or later slices. It shifts the distribution from the author's own writing to the terms in the works they cite. For each citing publication, we resolve its reference list against the corpus, and where a cited work has usable title or abstract text, we draw its terms into the profile (see Figure~\ref{fig:method-ref}). We report two variants. The inclusive variant keeps every usable cited work and is the one reported throughout. The external-only variant drops the target's own works from the cited lists, so that cited vocabulary cannot recycle the author's own wording, and serves as a check. The slice is the year of the citing publication, so a citation is read as an act of the moment of citation: the vocabulary of an old paper cited in 1970 enters the 1970 profile. Two levels of weighting prevent the document size from taking over. Each citing publication contributes the same total mass, its usable cited works share that mass equally, and within a cited work the terms are weighted by their relative frequency, so that neither a long bibliography nor a long cited abstract dominates. Writing \(c^{\mathrm{ref}}_{T,s}(t)\) for the resulting weight of term \(t\) across the target's citing publications in slice \(s\), the profile is again a share,
\[
P_T^{\mathrm{ref}}(t \mid s)
= \frac{c^{\mathrm{ref}}_{T,s}(t)}{\sum_{u \in V_{\mathrm{ref}}} c^{\mathrm{ref}}_{T,s}(u)} ,
\]
and the field distribution \(P_F^{\mathrm{ref}}(t \mid \tau)\) follows the same construction.

\subsubsection{Citation Identity}

\begin{figure}[ht!]
\begin{center}
\includegraphics[scale=\methodschemascale]{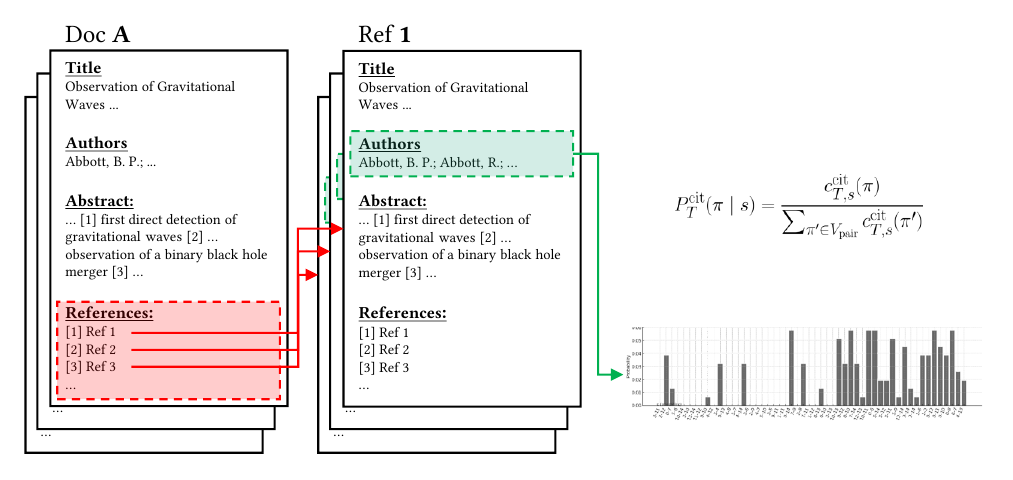}
\end{center}
\caption{Citation Identity, the same worked example at the relational layer. The pairs of authors the same target publication cites together form its co-reference profile \(P_T^{\mathrm{cit}}(\pi \mid s)\), a distribution over author pairs.}
\label{fig:method-cit}
\end{figure}

\textit{Citation Identity} asks which authorities and traditions an author binds together in their references, measured against the pairings the field forms. Works cited together share an intellectual relation \citep{small_co-citation_1973}, a relation scientometrics has long used to trace the structure of fields \citep{chen_structure_2010}. Following White's distinction between authors as citers and authors as cited objects \citep{white_authors_2001}, we read Citation Identity as the outgoing side, i.e. which authors an author cites and (consciously or unconsciously) combines. The feature is an unordered pair of cited authors, for example Einstein--Hilbert. Each citing publication contributes the same total mass, divided equally over the distinct pairs it forms, so that a long reference list does not dominate through the many pairs it can generate. Writing \(c^{\mathrm{cit}}_{T,s}(\pi)\) for the weight of pair \(\pi\) across the target's publications in slice \(s\), the profile is
\[
P_T^{\mathrm{cit}}(\pi \mid s)
= \frac{c^{\mathrm{cit}}_{T,s}(\pi)}{\sum_{\pi' \in V_{\mathrm{pair}}} c^{\mathrm{cit}}_{T,s}(\pi')} ,
\]
with \(V_{\mathrm{pair}}\) the set of pairs, and the field distribution \(P_F^{\mathrm{cit}}(\pi \mid \tau)\) follows the same form. Figure~\ref{fig:method-cit} visualizes the construction from a reference list.

This measure records a field's constellations of authority. Joining Einstein with Mach in a reference list points to a different lineage than joining Einstein with Wheeler or with Penrose, and an author who returns to the same pairings binds those authorities together repeatedly. We drop same-author pairs, since one repeatedly cited authority is anchoring rather than pairing, and we drop every pair that involves a cited work by the target, from the target's profile and the field baseline alike. Otherwise the baseline would contain the field's citing of the target, and an author would partly be compared against their own echo. This keeps outgoing reference practice distinct from the incoming reception defined below. Diagnostics and counting variants are given in the Supplementary Material.

Citation Identity is a relational dimension and therefore differs from the previous linguistic measures. An author can share the field's vocabulary and even its cited literature while binding authorities together in their own way, or do the reverse. In the canonical implementation, pairs are formed from the first listed author of each cited work and counted document-fractionally. The robustness boundaries below show how far individual readings depend on those choices rather than on a broader notion of an author's citation identity.

\subsubsection{Embedding Density Estimation}

\begin{figure}[ht!]
\begin{center}
\includegraphics[scale=\methodschemascale,trim=50pt 0pt 0pt 0pt,clip]{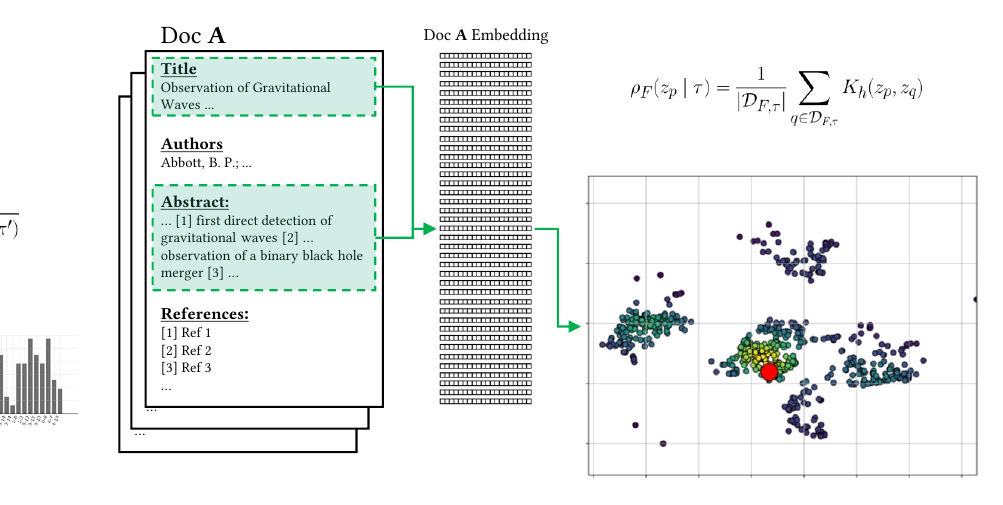}
\end{center}
\caption{Embedding Density Estimation, the same worked example at the level of document embeddings. The same target publication is embedded from its title and abstract, and the field density \(\rho_F(z_p \mid \tau)\) at its embedding coordinate is estimated from the field publications in the corresponding slice. A lower negative-log value marks a denser neighbourhood.}
\label{fig:method-ede}
\end{figure}

\textit{Embedding Density Estimation} (EDE) asks how densely the field occupies the semantic neighbourhood of an author's work, and how that occupancy changes over time. It keeps the target-field comparison but changes the object. Where the three measures above compare distributions over discrete features, EDE compares positions in a continuous semantic space. Each title + abstract is embedded as a numerical vector, so publications with similar content are closer together \citep{reimers_sentence-bert_2019}. For a target publication \(p\) with embedding coordinate \(z_p\), and field publications \(q\) in slice \(\tau\) collected in the set \(\mathcal{D}_{F,\tau}\), the field density at \(z_p\) is
\[
\rho_F(z_p \mid \tau)
= \frac{1}{|\mathcal{D}_{F,\tau}|} \sum_{q \in \mathcal{D}_{F,\tau}} K_h(z_p - z_q) ,
\]
where \(K_h\) is a Gaussian kernel of bandwidth \(h\). Before estimating the density we standardise the embedding coordinates over the whole analysed corpus, so that one slice's density is read on the same scale as another's, and we set \(h\) by Scott's rule \citep{scott_multivariate_1992} in the standardised space, \(h = n^{-1/(k+4)}\) for \(k\) coordinate dimensions, with \(k=2\) in the primary layer and \(n\) the number of analysed publications in the whole corpus. The bandwidth is computed once on the standardised whole-corpus coordinates and then held fixed across slices, so slice size does not change it and no separate baseline is fitted per period. Across the target's publications in slice \(s\), the slice score is the median negative log density,
\[
\mathrm{EDE}(s \mid \tau) = \operatorname{median}_{p}\big[-\log \rho_F(z_p \mid \tau)\big] ,
\]
so a lower score marks publications in a crowded region of the field and a higher score a sparser one. Unlike the earlier study's raw density, where higher meant denser, we report negative log density here, which makes the value distance-like, since every density slope and lead/lag sign in this paper follows that convention. A positive density slope is then movement toward sparser neighbourhoods over time, and a positive density lead/lag means a target slice is closest to a later field state. Figure~\ref{fig:method-ede} shows a target publication against the field in the embedding space.

EDE can register semantic proximity that exact-term and pair distributions may miss. Two papers can share almost no exact terms and still sit close together if they treat the same underlying problem, so density follows proximity in content, not agreement in vocabulary. A paper on gravitational collapse can lie in a sparse neighbourhood in one decade and a crowded one in the next if the field concentrates there, without any single term having to match. 

The quantity is field density at the embedded target positions, not position on a map. A neighbourhood at the visual edge of a two-dimensional projection can be densely populated, and a central one can thin out if the field grows faster elsewhere. Because the density is normalised, the field growing as a whole does not by itself raise it. Density estimation in the full embedding space is unreliable, because nearest-neighbour distances flatten as dimension grows and the bandwidth then governs the result more than the data, so we estimate density in a low-dimensional projection \citep{wang_understanding_2021}. The values depend on the chosen embedding model and coordinate representation. We treat 2D density created with the projection of \texttt{qwen3-embedding-8b}\footnote{qwen/qwen3-embedding-8b via \href{https://openrouter.ai/qwen/qwen3-embedding-8b}{OpenRouter}} as the primary layer and report Qwen's 5D and and additional 2D density created with \texttt{gemini-embedding-001}\footnote{google/gemini-embedding-001 via \href{https://openrouter.ai/google/gemini-embedding-001}{OpenRouter}} as robustness evidence, the 5D layer testing whether the choice of a two-dimensional projection drives the result. The target's own publications are excluded from the field whose density is estimated, so the score is the field's occupancy of the neighbourhood around the target rather than the self-density of the target's own cluster. The exact embedding-model identifiers, the title and abstract input with its translation handling, and the projection and standardisation are specified in the Supplementary Material. Density is independent from exact wording, such that there are no discrete features for the Welch/FDR layer and no term-by-term reading, and its interpretation leans on the other layers for lexical content.

\subsubsection{Citation Image}
\label{sec:citation-image}

\textit{Citation Image} asks how the field later receives an author, by how often and in what company it cites the target after publication. The term follows \citet[p.~328]{white_visualizing_1998}. Where the other measures describe an author's own practice, Citation Image describes the field's response to it, and the two have to be kept apart. We therefore introduce \textit{Citation Image} for interpretative support only and not as another target vs field measure. Reception is the channel through which later importance can distort the description of earlier work, since an author who was heavily cited in the 1990s was not necessarily for that reason central in the 1950s. A description that folded citation counts into the measures of practice would turn the most cited authors into the field's most representative ones by construction, which is retrospective canonisation. Reporting Citation Image as a separate measure, and correcting it for the visibilities built into the citation data, tries to prevent this. Four corrections are applied, each against a specific distortion. The basic count, \textit{field-internal future no-self reception}, includes only citations from later GRG field publications and excludes the target's own citing works, so that self-citation does not inflate reception. \textit{Exposure-normalised reception} divides this count by the years of exposure, so that long-published authors are not favoured by time alone. \textit{Coauthor-fractional citation counting} splits each citation event across the authors of the cited work, so that a heavily coauthored paper does not count in full for every one of its authors. No normalisation can recover reception outside the corpus window, so for authors who entered the field late, the reception recorded before the corpus ends around 2000 is read as a lower bound. In addition, received co-citation contexts record which other authors the field places beside the target. These contexts are read qualitatively in the Results. The exact weighting can again be found in the Supplementary Material.

The design choices shared by these measures, and the historical reading of each, are collected in one table in the Supplementary Material.

\subsection{Relating the measures}
\label{sec:relationmeasures}

For each author and measure we summarise the relation to the field as a level, as a slope over time, and, where it applies, as a lead/lag value. The level states how far an author typically diverges from the field, the slope whether that divergence grows or shrinks over time, and lead/lag which period's field state the author's features most resemble. We write \(d_i^m(s,\tau)\) for the target--field comparison of author \(i\) under measure \(m\), with the target in slice \(s\) and the field in slice \(\tau\). For the three distributional measures \(d_i^m\) is the divergence \(D_{\mathrm{KL}}(\widetilde P_T \parallel \widetilde P_F)\) defined above, and for Embedding Density it is the median negative-log density \(\mathrm{EDE}(s\mid\tau)\). The synchronous value sets \(\tau=s\),
\[
x_i^m(s) = d_i^m(s,s),
\]
and the three summaries are built from it. The level is its median over the author's active slices, \(L_i^m = \operatorname{median}_s x_i^m(s)\). The slope is the ordinary-least-squares trend of \(x_i^m(s)\) on slice time, \(\beta_i^m = \operatorname{slope}_s x_i^m(s)\), and is left undefined for fewer than two active slices. The lead/lag first finds, for each target slice, the field slice the target most resembles, \(\tau_i^{m\ast}(s)=\arg\min_{\tau} d_i^m(s,\tau)\), and records the signed gap \(\delta_i^m(s)=\tau_i^{m\ast}(s)-s\) in years. The reported author-level value is the unweighted mean of these gaps over the active slices,
\[
\Delta_i^m = \operatorname{mean}_s\, \delta_i^m(s),
\]
and the level reported alongside it is the mean of the divergences at those best-matching field slices, \(\operatorname{mean}_s d_i^m\big(s,\tau_i^{m\ast}(s)\big)\). The field slice \(\tau\) ranges over every slice that clears the support thresholds for that measure, across the full 1911--2000 span, so the same window is searched for all four measures. A target or field slice enters a summary only when it clears those thresholds, so slope and lead/lag are read only for authors with enough active slices.

Each measure operationalises one facet of an author's participation under a fixed representation. Own Vocabulary is lexical, Referenced Vocabulary cited-lexical, Citation Identity relational, and Embedding Density Estimation semantic-spatial, and the only quantity each reports is comparative resemblance to the field at a chosen period. Own Vocabulary can be seen as the baseline dimension. An author can write in the field's current language while citing older literature, or hold a vocabulary of their own while working in a crowded part of the field. Splits like these between Own Vocabulary and the other measures are what the analysis looks for. The measure speaks only to the language an author writes in, and carries no claim about whom they cite or how they are cited. If a paper cites Einstein and Hilbert together, Referenced Vocabulary records the terms inside the cited Einstein and Hilbert papers, while Citation Identity records the pairing of the two names. The cited horizon is a dimension of its own, close to Own Vocabulary in form but free to move against it. 

Calculating level, slope, and lead/lag of these measures therefore answers different questions. We treat two measures as describing the same trajectory only when they agree on level, slope, and on the field period a slice most resembles, and when that agreement survives the robustness variants. No pair of the four measures meets that standard across the fifty authors, which is why we keep them as separate layers. We use Spearman rank correlations across authors for levels, slopes, and lead/lag values, and read them descriptively. We call a correlation close at \(\rho \geq 0.6\), moderate between \(0.4\) and \(0.6\), loose between \(0.2\) and \(0.4\), and near-independent below \(0.2\). These bands fix the wording used for the relations among measures throughout the Results. Two of the couplings also have a mechanical floor. Own Vocabulary and Referenced Vocabulary are both built from terms, and Referenced Vocabulary and Citation Identity both start from the same reference list, so each pair shares an input before any author is measured. A possible situation would be: In the 1920s the field cites Einstein and Hilbert together for their papers on the field equations, again and again. In the 1980s a single author x cites the two names together once more, this time for papers of theirs that had never been cited together before. Citation Identity sees the same old pairing Einstein--Hilbert in both cases and places the later author x near an early field state. Referenced Vocabulary sees a combination of terms the field had never made and places the same paper possibly near a later one. Agreement within these pairs is therefore expected only in part, and we discount close correlations between them accordingly. With fifty authors these correlations are point estimates, so a value close to a band boundary should be read as indicative.

\subsubsection{Lead/lag readings}
\label{sec:warrant}

Under a single measure, a lead/lag value places one representation of a target nearer to a later or an earlier field state. We read these two situations as \textit{later-field} and \textit{earlier-field resemblance}, and, where brevity helps, \textit{future-oriented} and \textit{past-oriented} serve as shorthand for the same two situations. Agreement among the measures licenses historical inference by degree. A single lead/lag value, positive or negative, is weak evidence, and nothing follows from it about the other representations. Two agreeing values from related inputs, Own and Referenced Vocabulary or Referenced Vocabulary and Citation Identity, are moderate evidence, discounted by the shared input described in Section~\ref{sec:relationmeasures}, which let each pair agree partly by construction. Three or four values that agree across the distinct representations mark a stronger tendency. When that agreement meets interpretable driving terms or co-cited pairs and what is independently known of the author, the case supports a bounded historical reading of how the author's trajectory relates to the changing field. Influence, priority, or anticipation lie beyond its scale, since they require evidence the measures do not carry, from reception, citation direction, archives, or historiography. But they hint at situations worthy of inspection or interpretation. The Results read agreement within this interpretative framework rather than turning any single value into evidence. A resemblance to earlier or later field states is a similarity between publication traces, not necessarily a claim of progress or backwardness. This holds for every case below, so the readings state what a profile shows rather than defend against what it does not.

\section{Results}

Applying the above methods to the fifty identities with the most corpus publications (Top-50, see Supplementary Material) does more than track how individual trajectories reflect the renaissance or more generally the development of the field. The development is itself visible via trajectories of this kind, and because the field baseline is reconstructed from the same corpus, each author is read against a field that all other authors, the remaining forty-nine among them, help to compose. Whether written vocabulary moves with cited vocabulary, whether Citation Identity adds anything beyond cited vocabulary, and whether Embedding Density Estimation agrees with them in slope, in lead/lag, or in neither is read from the correlations in Table~\ref{tab:slope-correlations} and the scatter plots of Section~\ref{sec:overview}. Whether incoming Citation Image behaves separately from outgoing Citation Identity is read in Section~\ref{sec:citimage}. All these differences can occur both between authors and within a single trajectory.

\subsection{Systematic overview: one field, several trajectory dimensions}
\label{sec:overview}

\begin{figure}[p]
\begin{center}
\includegraphics[width=\productionfigurewidth,height=0.81\textheight,keepaspectratio]{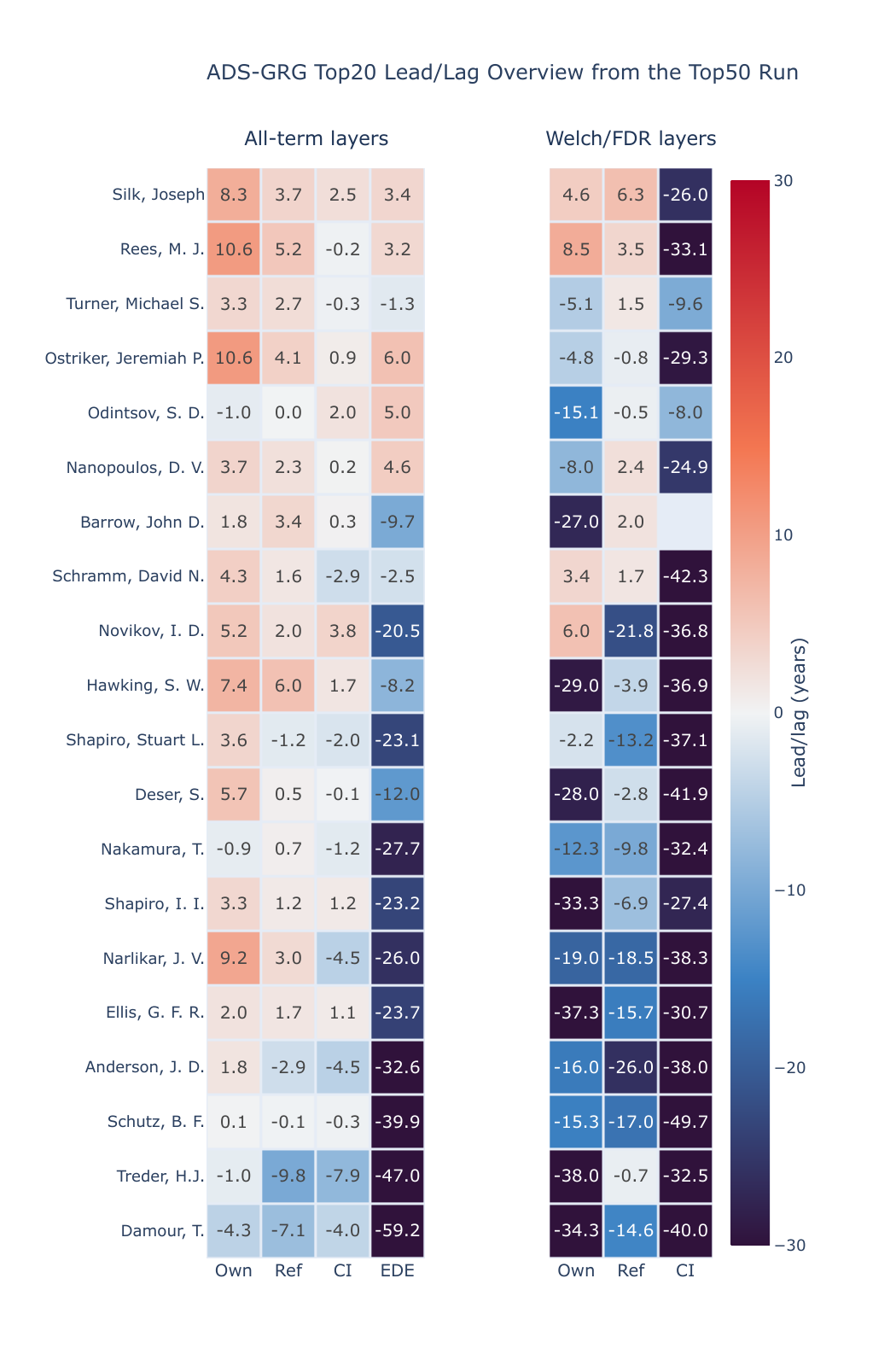}
\end{center}
\caption{Temporal field-state resemblance for the twenty most strongly covered of the Top-50 authors, shown as an overview. The full Top-50 values underlie the correlation tables. Positive values mark slices whose nearest field state lies later in time, and negative values mark slices closest to older field states. The left block compares each author with the full distributions for Own Vocabulary, inclusive Referenced Vocabulary, Citation Identity, and EDE. The right block keeps only the document-stable distinguishing vocabulary or reference-context features that pass Welch tests with Benjamini--Hochberg false-discovery-rate control at \(q\le0.20\), the exploratory candidate-feature layer of Section~\ref{sec:divergence}. Blank cells mark authors without usable feature support. The contrast shows that broad distributional resemblance and stable distinguishing features can point in different temporal directions.}
\label{fig:system-leadlag-all-terms}
\end{figure}

Figure~\ref{fig:system-leadlag-all-terms} gives a system-level view and shows that the four measures share structure in more than one dimension. We understand an author as close to the contemporary field when the author-level mean lead/lag lies within one year, i.e. half a two year slice step, and as later or earlier beyond that. In the all-term layer, Own Vocabulary places 38 authors closer to later field states, 9 near their contemporary field, and 3 closer to earlier states. Inclusive Referenced Vocabulary gives 30, 11, and 9, and the external-only variant gives 31, 10, and 9. Citation Identity is more balanced, at 10 later, 23 near, and 17 earlier. EDE runs against the broad vocabulary pattern, placing 15 authors closer to later field states, 2 near the contemporary field, and 33 closer to earlier field states. As stated above historians have shown that from the 1960s onward the GRG field turned more towards astrophysical and cosmological questions \citep{eisenstaedt_low_1989,blum_reinvention_2015,lalli_dynamics_2020,blum_gravitational_2018}. What the trajectories add is the anatomy of that turn in the authors who carried it.  The same publications point to different field periods depending on which layer is read. The language these authors write in and the language they cite resemble later field states, the semantic neighbourhoods those publications occupy lie where the field's mass was concentrated earlier, and outgoing co-citation stays close to the authors' own period. The turn these authors register thus runs as lexical renewal over the region that the small early field had filled and the later, larger field kept as one part among more. What the historiography describes as a shift of subject matter appears in the trajectories as a shift in how work is named and cited, while the field's share around that work follows its own course, which is why the same authors can count as future-oriented under one measure and past-oriented under another. The past lean of EDE is one aggregate result. It holds across embedding models, dimensions, and smoothing weights (Section~\ref{sec:robustness}), but it is also a property of the measure, which compares the share of each field slice near an author's publications, so the small, concentrated slices of the early field weigh heavily (Section~\ref{sec:limitations}). 

On the slope summary (\(\rho_{sl}\) in Table~\ref{tab:slope-correlations}), which records how fast a target moves toward or away from its contemporary field, the two vocabulary-based measures couple most closely (Own and inclusive Referenced Vocabulary, \(\rho_{sl}=0.737\)), Referenced Vocabulary tracks Citation Identity (\(\rho_{sl}=0.644\)), and density is the least correlated layer (\(\rho_{sl}\le0.32\) with each of the others). On the lead/lag summary (\(\rho_{ll}\) in Table~\ref{tab:slope-correlations}), which measures the field period a slice most resembles, the pattern shifts. EDE couples with the other layers (up to \(\rho_{ll}=0.562\) with Referenced Vocabulary) while the correlation between the two vocabulary measures relaxes to (\(\rho_{ll}=0.626\)). This is consistent with the opposed counts above, since a rank correlation records how the authors stand against one another and not how far a layer as a whole leans. Authors whose vocabulary points to later field states than their peers' also occupy semantic neighbourhoods that point to later field states than their peers', with the whole density layer displaced toward earlier field states. The density layer is therefore not a delayed copy of the lexical one. It orders the authors similarly as vocabulary does while sitting elsewhere in the field's history, so the offset belongs to the layer and not to particular authors. 

The per-author orientation classes order the pairs the same way. Own and Referenced Vocabulary fall in the same lead/lag orientation class for 33 of the fifty authors, Referenced Vocabulary and Citation Identity for only 22. Splitting the act of citing into a cited-language horizon and a configuration of authorities is therefore an empirical result and not only a design choice. Which literature an author draws vocabulary from and which authorities that author binds together are two decisions, and they can be made against different periods of the field. The two fall in different lead/lag orientation classes for 28 of fifty authors although both start from the same reference lists. These are the authors who, like the reader of Einstein and Hilbert in Section~\ref{sec:relationmeasures}, bind old authorities together under a vocabulary the field made later, or the reverse. Historical close reading has to focus on them. 

\begin{table}[ht!]
\caption{Spearman rank correlations between ADS-GRG Top-50 all-term author-level slope (\(\rho_{sl}\)) and lead/lag (\(\rho_{ll}\)) estimates in the main analysis. No multiple-testing correction is applied to this exploratory correlation table. The correlations are read descriptively.}
\label{tab:slope-correlations}
\begin{center}
\begin{tabular}{lrr}
\hline
Measure pair & \(\rho_{sl}\) & \(\rho_{ll}\) \\
\hline
Own Vocabulary--Referenced Vocabulary & 0.737 & 0.626 \\
Referenced Vocabulary--Citation Identity  & 0.644 & 0.442\\
Own Vocabulary--EDE & 0.312 & 0.492\\
Own Vocabulary--Citation Identity  & 0.304 & 0.265 \\
Citation Identity--EDE  & 0.237 & 0.375\\
Referenced Vocabulary--EDE  & 0.159 & 0.562\\
\hline
\end{tabular}
\end{center}
\end{table}

The Welch/FDR block in the right panel of Figure~\ref{fig:system-leadlag-all-terms} adds the document-level layer of Section~\ref{sec:divergence}. While the all-term comparison asks which full distribution is closest, Welch/FDR points to the document-stable distinguishing features. Under the \(q \le 0.20\) view, Own Vocabulary has 9 future-oriented authors, 1 near-field author, 39 past-oriented authors, and one author without usable feature support. Referenced Vocabulary is similar. The inclusive policy gives 15 future, 4 near, and 31 past, and external-only gives 11 future, 3 near, and 35 past with one author without usable feature support. Citation Identity points the other way. Its document-stable co-citation pairs are uniformly past-oriented, at 49 past with one author without usable feature support, because the pairs that most stably distinguish an author are recurring pairs of older authorities. Under the stricter \(q \le 0.10\) the picture holds, with Own Vocabulary tightening to 7 future, 1 near, and 40 past and two authors without usable feature support, the Referenced-Vocabulary variants staying past-leaning, and Citation Identity remaining uniformly past. One regularity runs through all of it. In every measure with a Welch/FDR filter the filtered counts lean further to the past than the all-term counts. The likely reason is what survives the filter. A feature passes only when it separates an author's documents from the field's documents consistently, and such persistent differences are habits kept while the field moved on, older terms retained and newer ones absent. An author who still writes of frozen stars in 1980 and not of black holes differs from the field of 1980 in exactly this stable term and hardly from the field of 1966, so the filtered comparison points back while the all-term comparison, carried by the vocabulary shared with 1980, does not. For co-citation pairs the effect is almost total, since a pair an author repeats is a fixed older pairing of authorities, White's stable core of recitations \citep{white_authors_2001}. The contrast is sharpest in Own Vocabulary and Citation Identity, where the broad distribution moves with the later field, or stays near-balanced, while the most stable terms and pairs stay older.

\begin{figure}[htbp]
\begin{center}
\includegraphics[width=\productionfigurewidth]{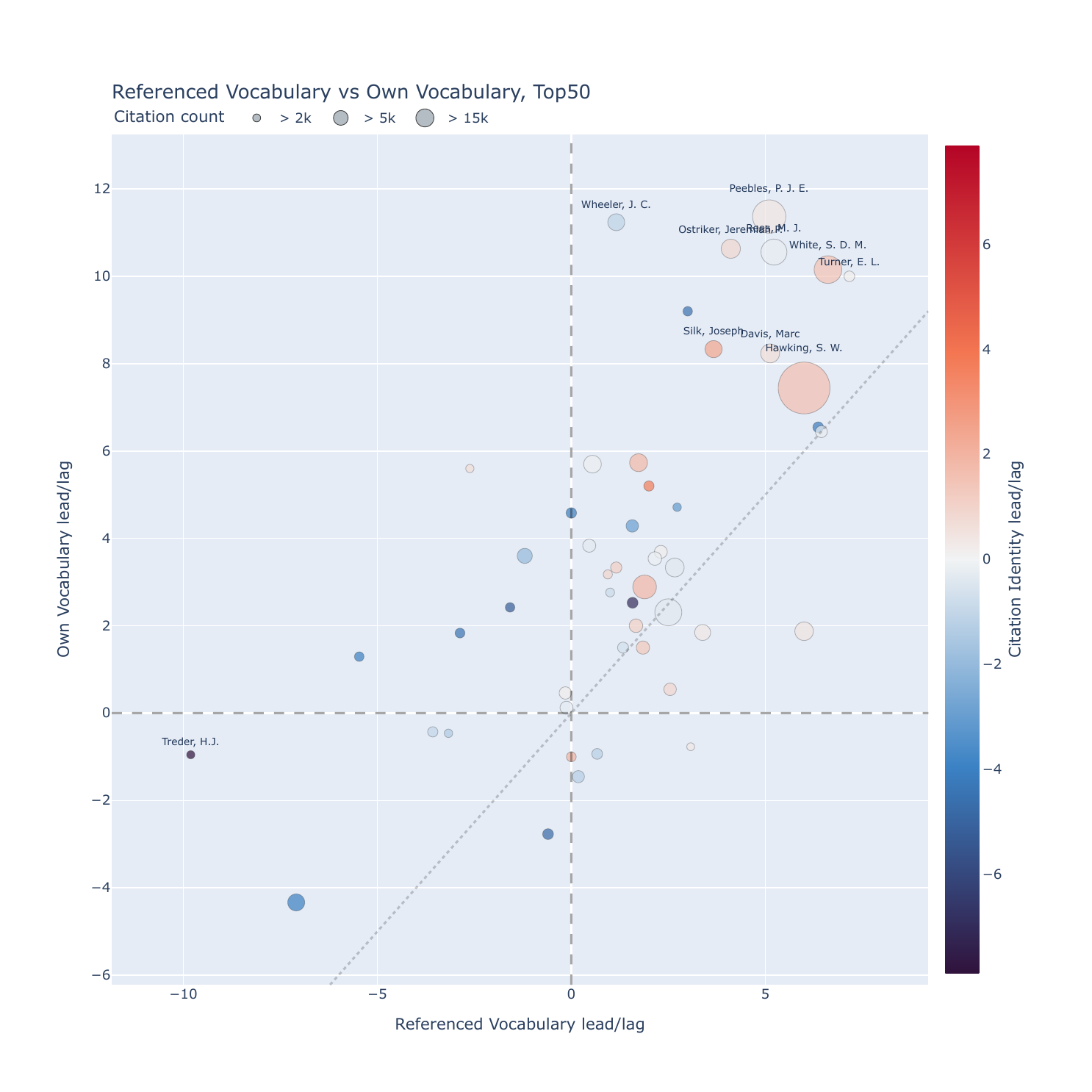}
\end{center}
\caption{Referenced Vocabulary and Own Vocabulary lead/lag. The dotted diagonal marks equal temporal orientation, and the dashed zero lines separate earlier-field from later-field matches. Point size follows citation count; colour shows Citation-Identity lead/lag. The close diagonal pattern shows that an author's own language and the language carried by cited works often move together, while cases away from the diagonal mark authors whose writing and cited literature occupy different temporal relations to the field.}
\label{fig:ref-vocab-own}
\end{figure}

\begin{figure}[htbp]
\begin{center}
\includegraphics[width=\productionfigurewidth]{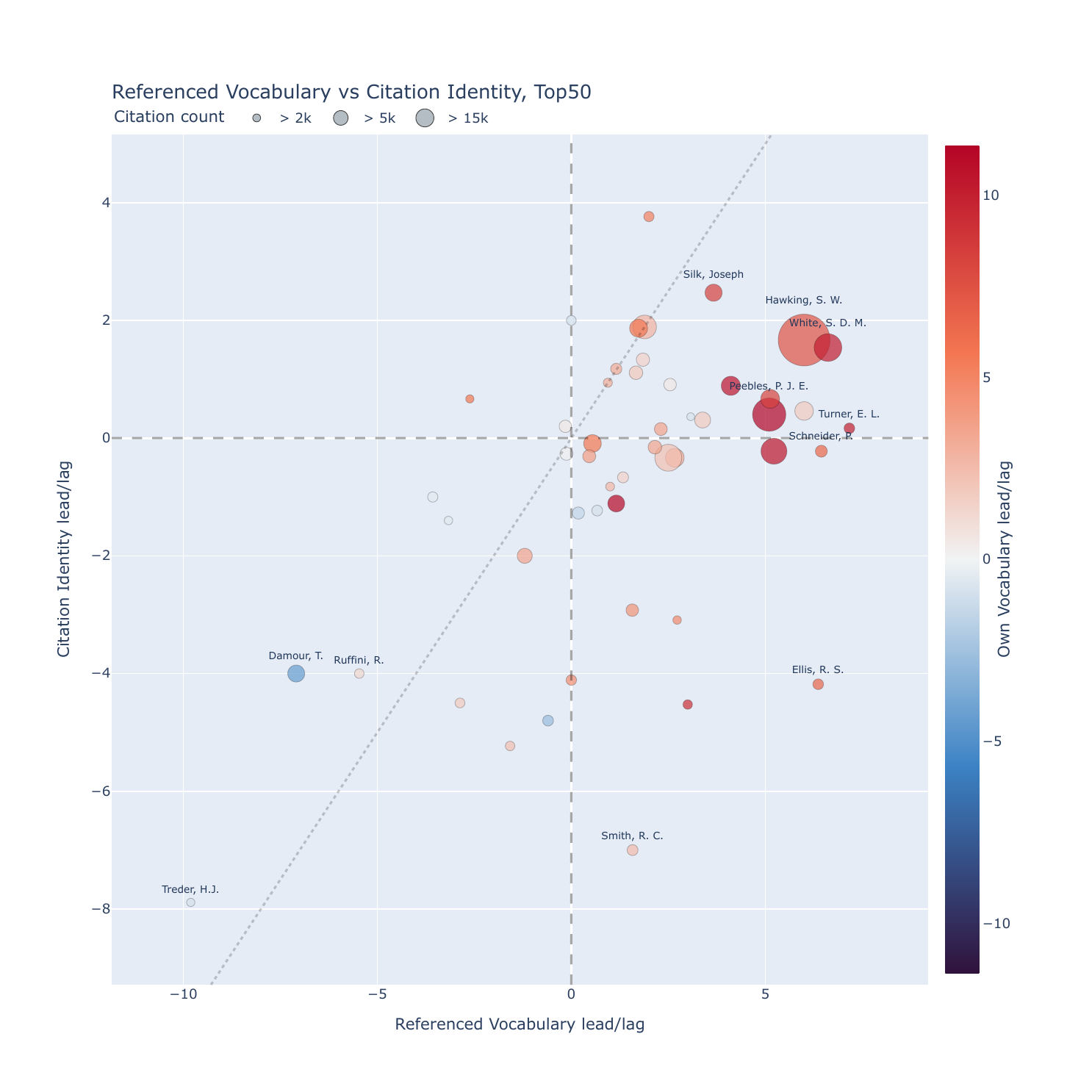}
\end{center}
\caption{Referenced Vocabulary and Citation Identity lead/lag. The horizontal axis follows the language of cited works; the vertical axis follows outgoing co-citation contexts. Point size follows citation count; colour shows Own Vocabulary lead/lag. Authors away from the diagonal show that citing literature with a later or older vocabulary is not the same as configuring later or older co-citation contexts.}
\label{fig:ref-vocab-citation-identity}
\end{figure}

\begin{figure}[htbp]
\begin{center}
\includegraphics[width=\productionfigurewidth]{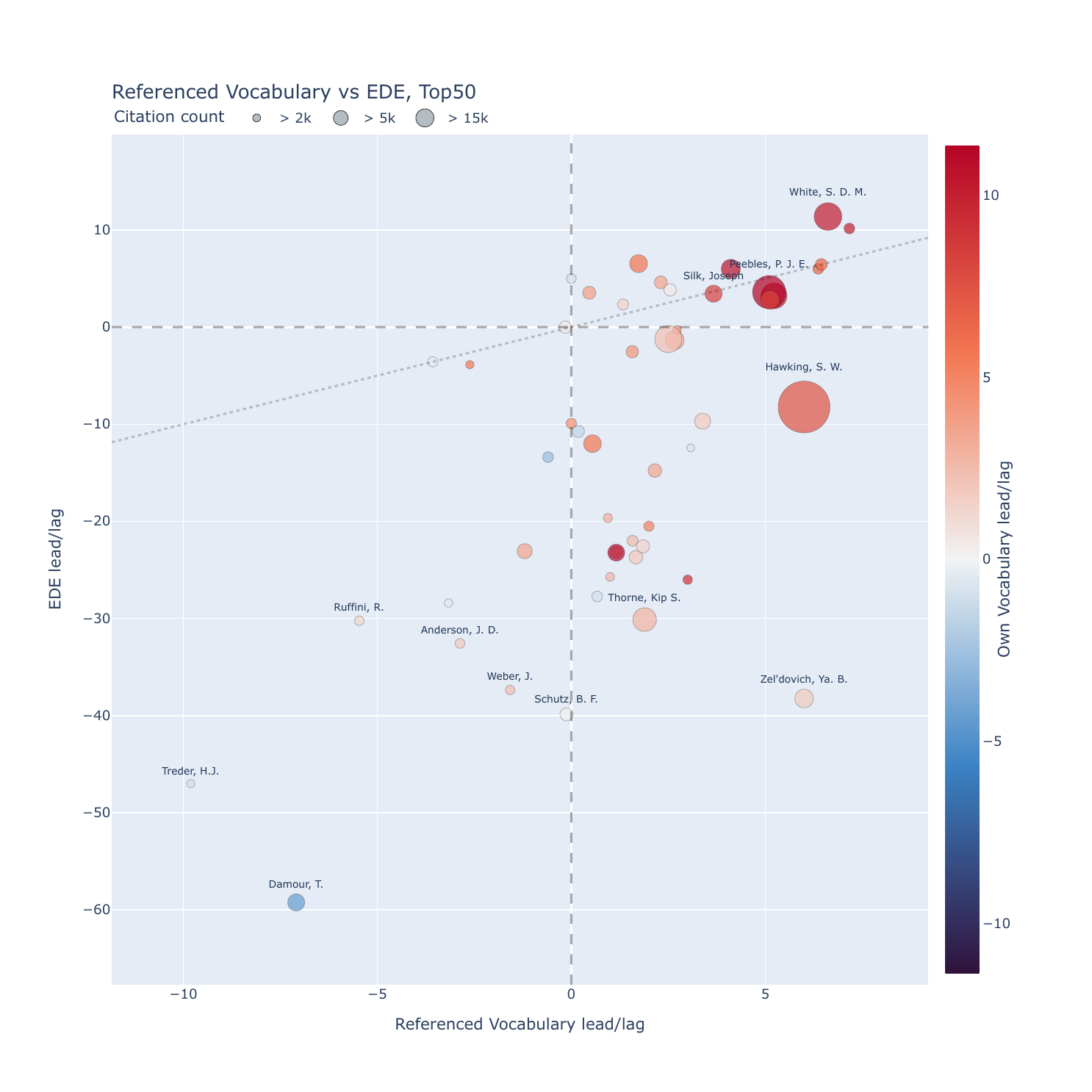}
\end{center}
\caption{Referenced Vocabulary and EDE lead/lag. The horizontal axis follows the language of cited works; the vertical axis follows the semantic neighbourhood around an author's publications in embedding space. Point size follows citation count; colour shows Own Vocabulary lead/lag. The wide EDE results visualizes the range of authors whose cited-language horizon is near or future-oriented while their semantic neighbourhood remains closer to older field states.}
\label{fig:ref-vocab-ede}
\end{figure}

Figures~\ref{fig:ref-vocab-own}--\ref{fig:ref-vocab-ede} make the same point by showing relations among the measures. Referenced Vocabulary sits close to Own Vocabulary and relates to Citation Identity, yet neither relation is identity (Figures~\ref{fig:ref-vocab-own}--\ref{fig:ref-vocab-citation-identity}). Citing literature with a later or earlier vocabulary is one thing, and configuring later or earlier co-citation contexts is another. Figure~\ref{fig:ref-vocab-ede} shows the split. The cited-language horizon can point near or later while the semantic neighbourhood stays older, which is why the density relation needs its own scale. In line with section~\ref{sec:warrant}, the agreement of several representations across the Top-50 is third-level evidence, and it licenses the system-level reading directly.

The Top-50 comparison therefore revises the two-case expectation of our previous work. The trajectories share one temporal axis, strongest between the language an author uses and the language they cite, and visible again where both meet semantic density in lead/lag. That axis is crossed by a second distinction, between lexical and citation horizons on one side and the local semantic-density neighbourhood on the other. Citation Identity follows the cited vocabulary only in part, staying more relationally specific and less stable. Density parts from the vocabulary measures in pace and joins them in period, so slope and lead/lag are two temporal dimensions of a trajectory, the pace at which a trace moves against its contemporary field and the field period it most resembles. 

The system-level result is that author-level participation in the changing field is not reducible to one field-position axis. The traces are structured but only partially coupled, and the off-diagonal authors, where the traces break their expected coupling, are the cases close reading has to examine carefully.

\subsection{Individual cases: authors closest to later field states}

\begin{figure}[p]
\begin{center}
\includegraphics[width=\productionfigurewidth,height=0.84\textheight,keepaspectratio]{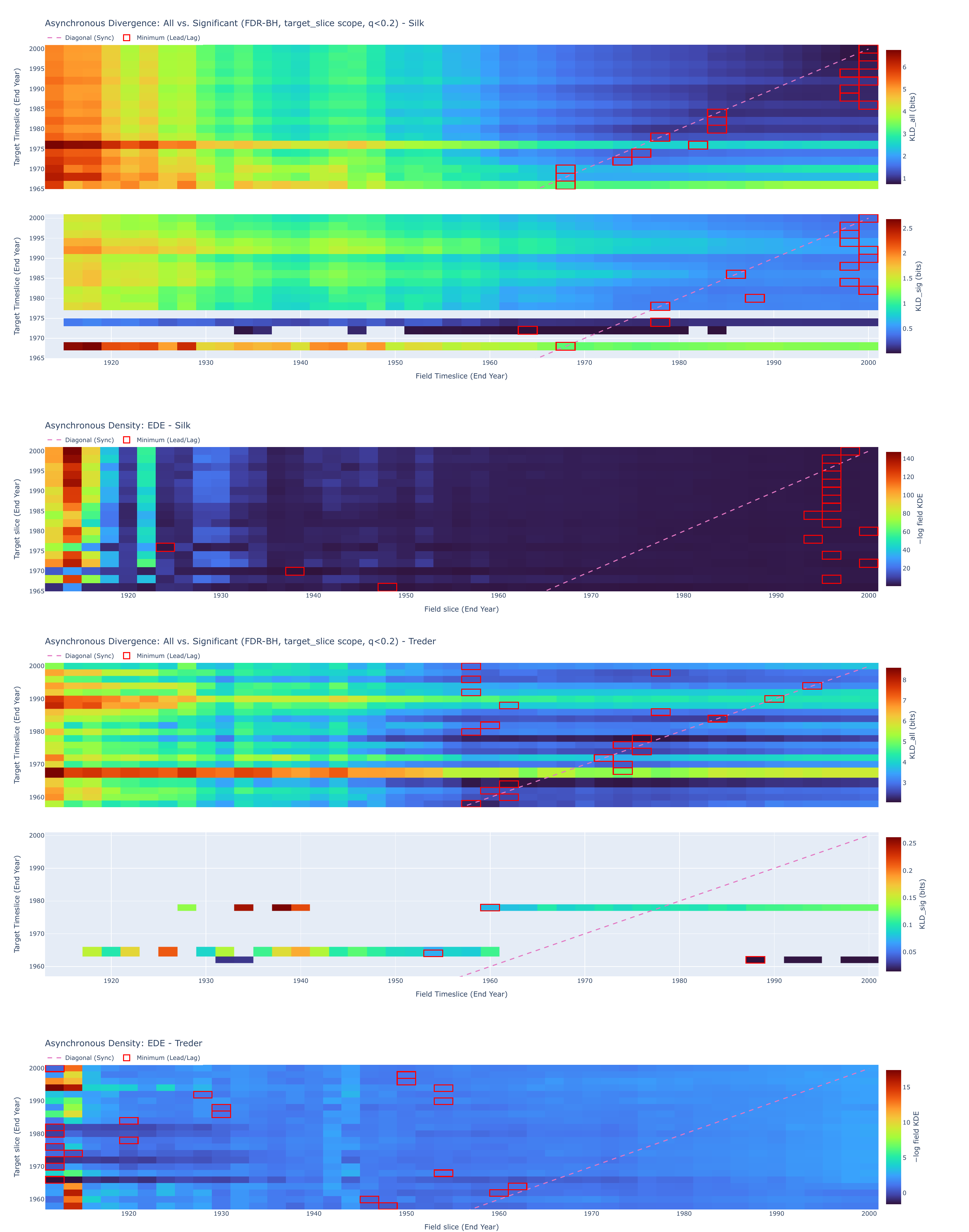}
\end{center}
\caption{Asynchronous trajectories of Silk (upper half) and Treder (lower half) in cited language and semantic density. For each author, the upper block compares the referenced vocabulary with earlier and later field slices, first over all terms and then over the document-stable Welch/FDR terms, where blank rows have no surviving terms, and the panel below it does the same for EDE. The pink diagonal marks the contemporary field slice, and red rectangles mark the closest field slice for each target slice. In Silk's later slices the minima frequently sit on or to the right of the diagonal, especially in the density panel, so his work increasingly resembles later GRG field states. Many of Treder's minima lie to the left of the diagonal, so his cited-language horizon and semantic neighbourhood repeatedly resemble older GRG field states.}
\label{fig:silk-dashboard}
\label{fig:treder-dashboard}
\end{figure}

\begin{figure}[htbp]
\begin{center}
\includegraphics[width=\productionfigurewidth]{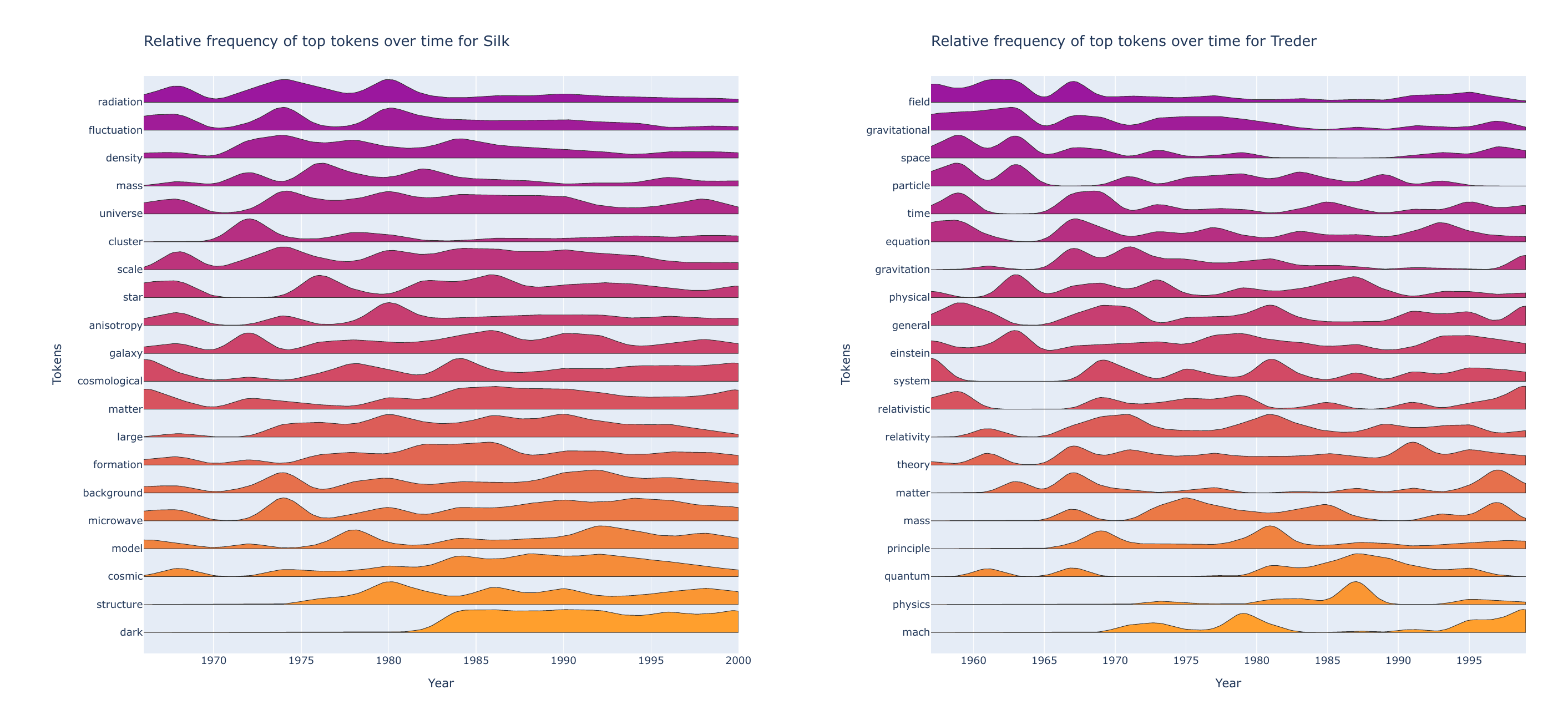}
\end{center}
\caption{Relative frequency of the top 20 Own Vocabulary terms for Silk (left) and Treder (right). Silk's vocabulary moves through radiation, fluctuation, density, mass, universe, galaxy, formation, background, microwave, cosmic structure, and dark matter. Treder's vocabulary remains organised around field, gravitational, equation, gravitation, Einstein, relativity, theory, principle, and quantum. The lexical contrast gives the historical content behind the lead/lag profiles: Silk's language belongs to an increasingly astrophysical and cosmological GRG, while Treder's retains a principle-oriented Einsteinian core.}
\label{fig:token-usage}
\end{figure}

The trajectory families in this section are interpretive groupings, without a formal cluster analysis behind them. The full Top-50 roster with coverage and reception per author is in the Supplementary Material, Figure~\ref{fig:system-leadlag-all-terms} shows the twenty most strongly covered of them. Silk's profile is the most direct bridge to the earlier two-case study, with Own Vocabulary (\(8.33\)), inclusive Referenced Vocabulary (\(3.67\)), Citation Identity (\(2.47\)), and Density (\(3.44\)) all pointing towards later field states on average. The upper half of Figure~\ref{fig:silk-dashboard} shows the two most relevant layers. The referenced-vocabulary minima often sit on or right of the synchronous diagonal, and his later density slices move toward later field states, though less strongly than the smaller comparison suggested. Figure~\ref{fig:token-usage} shows the lexical content as a development. The late-1960s slices already bind together primordial-fireball fluctuations, black-body radiation, and galaxy formation, and the later decades return to the same structure-formation questions through clusters, dark matter, and microwave-background anisotropy. Silk's later-field alignment is therefore legible within the maturing astrophysical and cosmological field trend, not a change of subject. The co-cited pairs run through the same problem space, joining Peebles with Bond, Smoot, and Davis, and the Own and Referenced Vocabulary slopes decline together as the field seems to move toward the contents Silk has been researching for some time.

Peebles moves from early physical-cosmology and galaxy-formation problems toward the clustering mathematics and parameter-era cosmology the later field concentrates on, his vocabulary furthest ahead in the set (\(11.37\)) while his co-citation stays near-synchronous (\(0.40\)) and gains distinctiveness. References are usable in 117 of his 163 publications, and Press--Efstathiou and Peacock--Efstathiou contribute most strongly to his Citation Identity, and both pairs are contemporary co-citations. The near-synchronous Citation Identity value therefore records reference configurations formed with the field of his own moment, while his written vocabulary runs well ahead of it. 

Where Peebles moves across problems, White's development is concentrated around simulation and halo structure. Cluster dynamics and terms around hierarchical halos appear in the late 1970s. Between 1980 and 1988, neutrino models, cold dark matter, N-body methods, filaments, voids, and dark halos turn his sub-corpus toward computational structure formation. The 1990s add COBE-era anisotropies, cluster normalisation, halo profiles, mock surveys, and public simulation data. Own Vocabulary (\(10.15\)), Referenced Vocabulary (\(6.62\)), Citation Identity (\(1.54\)), and EDE (\(11.38\)) all resemble later field states. White is the only case discussed here whose semantic neighbourhood points to later field states by as much as his vocabulary does, and the reason lies in where his papers sit. Simulation-based structure formation is a region the field filled only in the 1980s and 1990s, so the late slices are dense there and the early ones empty, while the papers of the relativists above lie where the early field had already gathered. His synchronous density falls across the career, from \(2.21\) in 1976 to \(1.69\) in 2000, placing his publications in progressively denser field neighbourhoods. 

Across Silk, Peebles, and White the leading outgoing co-citation pairings converge on the same small set of names. Peebles forms one half of the strongest pair episodes for both Silk and White, and Efstathiou for both Peebles and White. Their Citation Identity values stay at or near their own period (\(2.47\), \(0.40\), and \(1.54\)), far from the much earlier configurations of the next section. The authorities these authors most often bind together are contemporaries working on the same problems.

Linde's sub-corpus contributes mostly on discussions around inflation, which runs from field-theoretic cosmology through the 1982--1984 inflationary threshold to eternal inflation and preheating. His own and cited vocabularies (inflation, inflationary, scenario, chaotic) point towards the later field while Citation Identity and density stay near-synchronous. His declining vocabulary slopes record distinctive inflationary language becoming common field vocabulary. 

For Zel'dovich only 56 of 175 publications carry usable references, and English abstract coverage is limited, so the cited-literature chronology remains thin. Own Vocabulary is clearer. Terms around Relativistic cosmology and collapse precede a 1970--1972 hinge around gravitational instability, relic radiation, cluster gas, and quantum fields. Particle-cosmology and structure questions remain connected through the 1980s. Own and Referenced Vocabulary resemble later field states (\(1.88\) and \(6.00\)), Citation Identity is near-synchronous (\(0.46\)), but EDE points to much earlier field states (\(-38.24\)). Language characteristic of later GRG was used in Zel'dovich's Moscow programme of relativistic astrophysics, which moved from collapse and the hot universe in the 1960s to large-scale structure and the Sunyaev--Zel'dovich effect in the 1970s \citep{thorne_black_1994}, while its semantic neighbourhood resembles much earlier parts of the corpus.

The document-level layer keeps the reading from becoming too smooth. For Silk the stable vocabulary and cited-vocabulary candidates stay future-oriented (\(+4.6\) and \(+6.3\)) while his most stable co-citation pairs point firmly to the past (\(-26.0\)). This is later-field language configured over sedimented older authorities. Across the Top-50, Own Vocabulary's most stable distinguishing terms point mostly to earlier field states, the retained vocabulary of Section~\ref{sec:overview}, while the broad distribution moves with the field.

\subsection{Older horizons, infrastructure, and measurement cultures}

The strongest counter case remains Treder. His own language overall is near-synchronous with the field but conceals one of the clearest developments in the corpus. The layers built from his references and his semantic neighbourhood are consistently past-oriented, with Own Vocabulary \(=-0.95\), inclusive Referenced Vocabulary \(=-9.81\), Citation Identity \(=-7.89\), and Density \(=-47.00\). The lower half of Figure~\ref{fig:treder-dashboard} shows the repeated left-of-diagonal matches in Referenced Vocabulary and Density. His corpus moves through different phases rather than holding one vocabulary, from the technical relativity and unified-field problems of the late 1950s, through tetrad theory, teleparallelism, equivalence, post-Newtonian gravodynamics, and Mach--Einstein reorientation of the 1970s, to a late turn that carries those foundations into problems on cosmogony and Mach-versus-dark-matter arguments. The near-flat Own Vocabulary slope compresses a decline and later growing divergence based especially on usage of terms around Einstein, principle, and Mach. The durable Einstein-centred lexical core in Figure~\ref{fig:token-usage} is the most consistent trace and identifies a locally meaningful Einstein tradition, whose philosophical and institutional dimensions are independently documented \citep{schlattmann_relativity_2020}. His references reach the corpus thinly, 155 of 350 publications and under two references each, so the cited and relational layers rest on sparse coverage. The co-cited pairs name that older horizon directly, joining Dirac with Heisenberg and Weizs\"acker, Lorentz with Cunningham, and Papapetrou with Finkelstein rather than the astrophysicist/cosmologists who dominate the later field. The two bibliographic measures therefore require caution. Their earlier-field orientation nevertheless agrees with EDE. The metric profile thus documents from the mid 1970s onward a past orientation of own vocabulary, cited, relational, and semantic horizons under sparse ADS reference support. 

In the case of Damour a similar orientation has to be interpreted differently. His measures all point to earlier field states, with EDE especially strong at \(-59.23\). Within the corpus window his work moves from binary pulsars and the equations of motion of compact bodies in the 1970s, through post-Newtonian mechanics and the quadrupole-formula debates of the 1980s, to tensor--scalar gravity and, by the late 1990s, the effective-one-body and inspiral-template work that gravitational-wave modelling later was based on. His own terms mark this classical analytical relativity rather than a local tradition outside modern GRG. The corpus ends in 2000, before the detection era, so the gravitational-wave reading is a historically grounded post-window interpretation, which the past-pointing lead/lag values do not establish on their own \citep{kennefick_traveling_2007,blum_gravitational_2018}. All four slopes decline together, and the co-cited pairs hold the binary-pulsar and experimental-gravity tradition in view, joining Taylor with Thorne and Nordtvedt with Will. The field later receives this work through Blanchet, Taylor, Will, and Thorne, the community that turned the older analytics into detection infrastructure.

Thorne supplies an instrumental route through the same field transformation. His publications move from compact stars and relativistic gravitation into quantum measurement and gravitational radiation, then toward detector design, source modelling, and waveform problems associated with LIGO during the 1990s \citep{thorne_black_1994,kennefick_traveling_2007,blum_gravitational_2018}. Own Vocabulary, Referenced Vocabulary, and Citation Identity resemble later field states (\(2.89\), \(1.89\), and \(1.89\)), while EDE remains much earlier (\(-30.11\)). Among the outgoing pairs that contribute most are Estabrook--Rudolph, Einstein--Weber, Synge--Isaacson, and Braginskii--Giffard. Those pairs locate experimental and analytical reference configurations in Thorne's earlier outgoing references. The career-level split of the four measures makes detector and gravitational wave work visible as a distinct route through later GRG as part of the astrophysical turn.

Shapiro is the clearest single case of the split between measures this paper describes, and also a warning about Author Name Disambiguation. His Own Vocabulary point towards the later field through a measurement culture that runs from 1960s radar ranging and ephemeris tests of general relativity, through Mariner and Viking spacecraft ranging, geodesy, and gravitational lensing, with Own Vocabulary \(=3.33\) and Referenced Vocabulary \(=1.16\), while the semantic neighbourhood stays older at Density \(=-23.16\). The \textit{Shapiro, I. I.} identifier, however, seems to fold Irwin Shapiro's radar programme together with a quantum-gravity namesake whose titles enter the corpus from the mid-1980s. The measurement-culture reading therefore rests on the slices before that entry, and the Citation Identity value (\(1.18\)) is read only as a stability-flagged signal (see Section~\ref{sec:robustness} below), not as Irwin Shapiro's reference practice. The reception layer stays with the radar programme, placing Shapiro near Nordtvedt, Damour, Dicke, Brans, and Will, while the contaminated outgoing layers are flagged.

\begin{table}[ht!]
\caption{Author-level mean lead/lag values in years for the cases discussed in this section. Positive values indicate resemblance to later field states and negative values resemblance to earlier states. The EDE column ranges much more widely than the three KLD columns, since a semantic neighbourhood can match a field state several decades removed while a vocabulary rarely does. \textsuperscript{*}The \texttt{Shapiro, I. I.} row describes a mixed corpus identity and is not an Irwin-Shapiro-only profile.}
\label{tab:trajectory-families}
\begin{center}
\footnotesize
\begin{tabular}{lrrrr}
\hline
Author & Own Vocab. & Ref. Vocab. & Cit. Id. & EDE \\
\hline
Silk, Joseph & 8.33 & 3.67 & 2.47 & 3.44 \\
Peebles, P. J. E. & 11.37 & 5.10 & 0.40 & 3.60 \\
White, S. D. M. & 10.15 & 6.62 & 1.54 & 11.38 \\
Linde, A. D. & 2.31 & 2.50 & -0.33 & -1.23 \\
Zel'dovich, Ya. B. & 1.88 & 6.00 & 0.46 & -38.24 \\
Treder, H. J. & -0.95 & -9.81 & -7.89 & -47.00 \\
Damour, T. & -4.33 & -7.09 & -4.00 & -59.23 \\
Thorne, Kip S. & 2.89 & 1.89 & 1.89 & -30.11 \\
Shapiro, I. I.\textsuperscript{*} & 3.33 & 1.16 & 1.18 & -23.16 \\
Hawking, S. W. & 7.44 & 6.00 & 1.67 & -8.22 \\
\hline
\end{tabular}
\end{center}
\end{table}

\subsection{Citation Image and received contexts}
\label{sec:citimage}

\begin{figure}[htbp]
\begin{center}
\includegraphics[width=\productionfigurewidth]{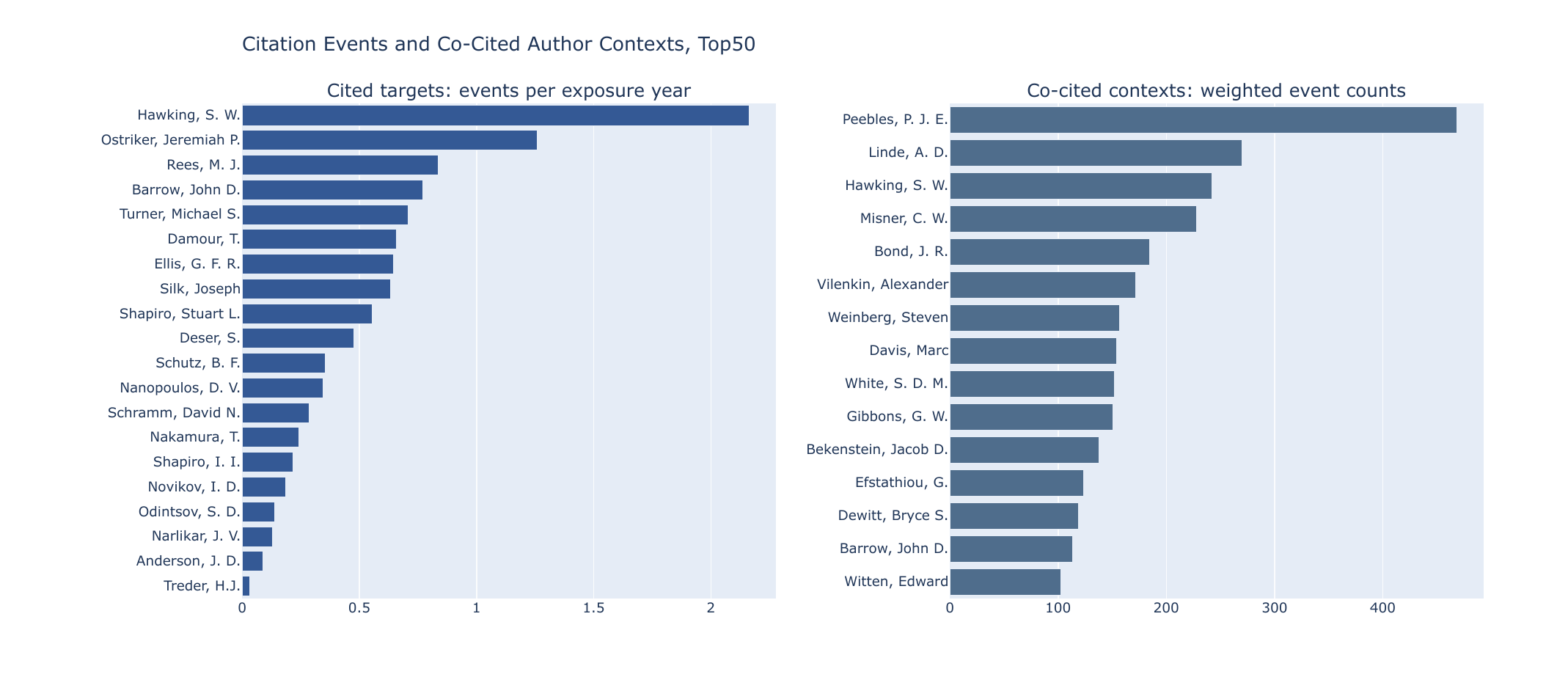}
\end{center}
\caption{Incoming reception (Citation Image) for the twenty most-received Top-50 authors, with the received co-citation contexts beside them. Left: targets ranked by field-internal, coauthor-fractional citation events per exposure year, self-citations excluded. Right: the fifteen authors most often cited alongside these twenty targets in the field papers that cite them, with weights summed across all twenty targets. The right panel therefore shows the shared reception context of the group.}
\label{fig:citation-image}
\end{figure}

Citation Image records a later selection made by the indexed field. The left panel of Figure~\ref{fig:citation-image} shows an exposure-normalised measure for the twenty most publication-rich Top-50 authors. It counts future field-internal citation events after removing self-citations and allocates each event fractionally across the cited work's coauthors. The right panel aggregates the authors that occur beside targets in papers citing them (that are co-cited with them). In the raw field-internal count reported in the Supplementary Material, which covers the full Top-50 rather than the twenty identities plotted here, Peebles has the most events (8,578), followed by Hawking (8,354). Hawking also has the largest global coauthor-fractional count.

Damour's no-self-citation output illustrates this attribution. Its highest-weight received-author contexts are Blanchet, Taylor, Will, and Thorne, placing his work within post-Newtonian mechanics, binary-pulsar research, and precision gravity. This is separate from the outgoing Taylor--Thorne and Nordtvedt--Will pair episodes in Damour's own references. The two directions together connect his analytical trajectory to the GRG community that later reused it.

The case of Hawking is an example for a split reception. His incoming reception is among the largest in the set (about 8400 field-internal events, second only to Peebles, and the largest global co-author-fractional count), placing him with Vilenkin, Bekenstein, Gibbons, Linde, and Coleman, yet the internal layers do not reduce to it. Own Vocabulary (\(7.44\)), Referenced Vocabulary (\(6.00\)), and Citation Identity (\(1.67\)) point toward the later field, while EDE stays older at \(-8.22\). The profile is a sequence of jumps rather than one of recurring vocabulary, running from singularities and expanding universes in the late 1960s, through black-hole mechanics and path-integral quantum cosmology, to pair creation, open inflation, and brane-world black holes by the late 1990s. Several target slices from 1966 to 1978 best match much later field periods, whereas 1980 and several slices from 1988 to 1994 jump to the earliest decades of the corpus. Those large negative gaps pull the career mean below zero. The EDE mean therefore condenses alternating re-embeddings rather than an older neighbourhood carried unchanged through the career. The outgoing co-citation runs through wormhole, information, and pair-creation pairs such as Coleman--Giddings and Klebanov--Coleman, a different structure from the reception the field later built around him. This example shows that public visibility is not necessarily technical reception.

\subsection{Robustness of results}
\label{sec:robustness}

We test the results against one changed design choice at a time, beginning with the document embedding and its dimension for density. Qwen 2D gives 15 future, 2 near, and 33 past authors, Qwen 5D gives 16, 2, and 32, Gemini 2D gives 15, 4, and 31. The rank relation is strong between Qwen 2D and Gemini 2D (\(\rho=0.914\)) and still substantial between Qwen 2D and Qwen 5D (\(\rho=0.773\)). This supports the aggregate claim that density is more past-oriented and more independent than the vocabulary layers, but it does not make every individual density value interchangeable across embedding models or dimensions.

For the measure of Citation Identity we gradually relax the canonical design choices one at a time (listed in the design-choice table of the Supplementary Material). We drop every pair involving a cited work by the target, in the target's own documents and in the field baseline alike. Applying that exclusion only inside the target's documents, so that the baseline still carries the field's citing of the target, leaves the author ordering essentially unchanged (\(\rho=0.994\), no slope sign reversals). Dropping the exclusion entirely does change it. The author is then compared partly against their own reception, rank agreement with the canonical design falls to (\(\rho=0.638\)), and six slope signs reverse. Binary and multiplicity counting preserve broad rank structure only partly and produce 17 to 22 lead/lag orientation changes. Moving from first-author to all-author cited-work expansion produces 24 lead/lag orientation changes. Citation Identity is therefore usable as first-author outgoing evidence, with individual claims qualified where coauthorship or author scope is historically central.

In the case of Referenced Vocabulary, removing the external-only policy changes levels for cases with substantial self-citation or school-specific recycling, but the aggregate orientation stays close to inclusive Referenced Vocabulary. External-only is therefore a check on leakage from an author's own writing into the cited-language horizon, and not a substitute for the main cited-literature measure.

Two further checks bound the smoothing and the anchor cases. Raising the smoothing weight from \(\lambda=0.05\) to \(\lambda=0.5\) leaves the density orientations unchanged and shifts the lead/lag orientation of nine authors in Own Vocabulary and ten in Citation Identity. The aggregate density pattern is thus the most smoothing-stable. Removing Silk and Treder and recomputing across the remaining forty-eight authors preserves the rank order of every slope and lead/lag correlation, with magnitudes shifting only modestly, so the coupling structure does not depend on the two reference-point authors.

The robustness checks support the paper's main claim that the measures overlap as correlated but distinct layers, while showing that no single lead/lag sign can be turned into a trajectory description without further context. They are rather a guide to which claims can be broad, which must be local, and which require closer historical follow-up.

\section{Discussion}

\subsection{Differentiating the renaissance}

Our Top-50 comparison revises the hypothesis behind the earlier two-case study \citep{schlattmann_trajectories_2024}. At fifty authors the single story of the two-case study becomes a special case. Vocabulary, cited-vocabulary, semantic neighbourhood, outgoing reference context, and incoming reception behave as separable relations. 

The separation among these traces differentiates at least two historical developments that are easily collapsed under the term \textit{renaissance}. The institutional and intellectual recovery of general relativity began during the 1950s and early 1960s, before the later concentration on quasars, background radiation, compact objects, and cosmological structure \citep{blum_renaissance_2020,blum_reinvention_2015,lalli_dynamics_2020}. Gravitational-radiation research also connected compact-object dynamics and precision measurement to source modelling and detector work, an experimental/instrument driven strand visible in Thorne's trajectory and in Damour's analytical one \citep{kennefick_traveling_2007,blum_gravitational_2018}. For most of the Top-50 authors the comparison begins after the earlier community had started to form and follows how researchers who entered at different moments and in different settings moved through these later reorganisations.

Own Vocabulary places 38 authors nearer later field states, and inclusive Referenced Vocabulary places 30 there. EDE places 33 nearer earlier states. Many trajectories therefore combine language associated with later GRG states with semantic neighbourhoods in which the field's share was largest in earlier periods.

Later-field resemblance is common in the all-term distributions, whereas filtered document-stable distinguishing terms and co-citation pairs usually resemble earlier field states. As cosmological and astrophysical vocabulary spread across GRG, some of it ceased to distinguish the authors who used it. Career-specific older terms and authority pairings nevertheless persisted within that shared language.

\subsection{The historical meanings of past and future}

Lead/lag, as introduced in Section~\ref{sec:warrant}, locates a resemblance and says nothing about its cause. It compares each trace with the field's own changing publication space, so the baseline already has a history. A match with a later field state may arise from terminology, a concentration of problems, or changing database coverage. An earlier-field match may follow from an inherited approach or a locally sustained programme. Weak indexing can produce a similar value. Publication chronology and historical evidence decide which of these applies.

Treder and Damour show why this distinction matters. Both have strongly earlier-field EDE and earlier cited horizons, yet their slopes describe opposing movements. Treder's Referenced Vocabulary and Citation Identity become more distinctive over time, while Damour's three KLD layers converge with the contemporary field. But the same earlier-field sign accompanies a more locally formed principle oriented programme in Treder and an evolving analytical resource in Damour.

Table~\ref{tab:typology} maps recurring evidential situations without assigning authors to stable types. Its rows may overlap within one career, and the relevant relation can change from one phase to the next.

\begin{table}[ht!]
\caption{A typology of layered field participation in the ADS-GRG Top-50. Membership can overlap. The examples are read from the lead/lag summaries in Table~\ref{tab:trajectory-families}, under the same near band and identity flags (\textsuperscript{*}, \textsuperscript{\dag}).}
\label{tab:typology}
\begin{center}
\footnotesize
\begin{tabular}{>{\raggedright\arraybackslash}p{0.18\textwidth}>{\raggedright\arraybackslash}p{0.22\textwidth}>{\raggedright\arraybackslash}p{0.27\textwidth}>{\raggedright\arraybackslash}p{0.18\textwidth}}
\hline
Pattern & Measure signature & Historical reading & Illustrative cases \\
\hline
Later-field integration & Own, Referenced, and density later; Citation Identity near-synchronous or later & The trace aligns with later field states across every layer, with co-citation formed alongside the field. & Silk, Peebles, White \\
Lexical future over an earlier field concentration & Own and Referenced later, density earlier & Later vocabulary and references at a location where the field's share peaked earlier. & Shapiro\textsuperscript{*}, Hawking \\
Older reference tradition & Referenced and Citation Identity earlier, Own near or later & A contemporary idiom over older cited and configured authorities. & Treder, Weber, Anderson\textsuperscript{\dag} \\
Infrastructure taken up after the window & Density strongly earlier despite later importance & Older formal machinery becomes central after the window. & Damour, Schutz \\
Reception split & Citation Image high, internal layers mixed & The field receives the author differently from their outgoing reference practice. & Hawking, Linde \\
\hline
\end{tabular}
\end{center}
\end{table}

The map also keeps data limitations outside the historical typology. Because the corpus ends in 2000, late target slices have fewer later field states available for comparison, and Citation Image cannot observe reception after 2000. This corpus-edge asymmetry supplies no historical meaning for either sign of a lead/lag value. Mixed identities can generate internally coherent numbers for a historically incoherent target, as the Shapiro case shows. Both conditions must be checked before a temporal resemblance is interpreted.

Citation Image introduces a second time axis, the later field's return to an author's publications. Hawking's high field-internal reception accompanies later lexical resemblance and an earlier-field EDE mean produced by a non-monotonic path. Treder combines the largest publication count among the targets with very low later GRG-internal reception. Historically, they record how unevenly ADS-indexed GRG returned to these careers and, through the received contexts, where it placed their work.

\subsection{Transfer to other research areas}

The approach applies to other research areas as a layered historical comparison. It needs a diachronic corpus with enough temporal depth, metadata linking documents to actors, a historically interpretable field boundary, and adequate text, embedding, and reference data. We applied it to two further datasets as an implementation check, an INSPIRE HEP holography corpus and a Semantic Scholar ACL corpus (Supplementary Material). The close coupling between written and cited vocabulary that organises GRG reappears in the holography corpus (\(\rho_{sl}=0.80\) against \(0.74\) for GRG) and loosens in the computational-linguistics corpus (\(\rho_{sl}=0.35\)), so the relation between the measures is itself a measurable property of a field as registered in its data.

\subsection{Limits and reproducibility}
\label{sec:limitations}

The main limitations are historical as much as technical. Titles and abstracts are incomplete representations of publications, especially older ones, and translation into English improves comparability while erasing some historically meaningful language differences. Referenced Vocabulary inherits this and adds uneven coverage of reference titles and abstracts, which weighs on Russian, German, Japanese, French, and other non-English bibliographies. The same unevenness runs through the cases above, from Treder's under-reported references to Zel'dovich's and Novikov's transliterated Soviet work and Nakamura's Japanese-school context, so ADS visibility is itself historically structured. Citation data is incomplete and unevenly distributed across languages, journals, and publication forms, and author disambiguation is imperfect\footnote{The clustering approach used here reaches an F1 score of 97.0\% on a labelled ADS sample; code and evaluation are available at \url{https://github.com/raphschlatt/ads-and}.}. The density lead/lag compares the share of each field slice near an author's publications, so it records the field's expansion away from a location as much as the location itself. A within-slice rank of density would remove the size effect and is reserved for a follow-up analysis.

This matters most for shared surnames, the two Shapiros, three Ellises, two Smiths, and the Wheelers in the target set, and for off-domain material inside a matched identity, such as Davis's pre-1978 rows and the supernova stream in \texttt{Wheeler, J.\,C.}, see Supplementary Material for an overview. The corpus boundary carries the same unevenness: the Top-50 includes Fritts, whose profile is past-oriented in three of four layers and belongs on inspection to atmospheric gravity-wave research that the keyword boundary admitted alongside gravitational-wave work. We keep the corpus baseline as the historically formed publication space and read as named GRG cases only those targets whose membership the historical record supports. Removing a single flagged boundary case such as Fritts shifts the orientation counts by at most one author and leaves the correlation structure intact.

\section{Conclusion}

This article turned the two-case comparison of \citet{schlattmann_trajectories_2024} into a systematic one and asked whether Own Vocabulary, Referenced Vocabulary, Citation Identity, and Embedding Density describe the same trajectory of an author through a changing field. Across the fifty most-published GRG authors, no pair meets the standard set in Section~\ref{sec:relationmeasures}. Written and cited vocabulary move most closely together and usually resemble later field states. Citation Identity keeps pace with the cited vocabulary but dates 28 of the fifty differently from the vocabulary drawn from the same reference lists. Density records where the field's publications were concentrated, early for most of the fifty, while the field diversified around them. An author's changing distance from the contemporary field and the historical field period their work most resembles consequently need separate interpretations.

In GRG the astrophysical turn that followed the renaissance appears most widely in language, since most of the fifty write and cite in the vocabulary of later, more astrophysical field states. Beneath that shared language, stably distinguishing terms resemble earlier field states for most of the fifty and repeated authority pairings for nearly all, so the problem traditions that the reinvention of the 1950s and early 1960s had connected \citep{blum_reinvention_2015,lalli_dynamics_2020,lalli_socio-epistemic_2020} persist beneath a change in the language written and cited. The measures also separate developments that the renaissance narrative tends to collapse. Silk's and White's research on structure formation resembles later field states across all four measures. Treder and Damour both resemble earlier field states in cited language and semantic neighbourhood, yet Treder's thinly indexed cited profiles become increasingly distinctive while Damour's analytical work converges with the contemporary field. The subsequent use of Damour's work in gravitational-wave modelling gives that earlier resemblance a historical significance that our corpus, ending in 2000, cannot establish by itself \citep{kennefick_traveling_2007,blum_gravitational_2018}.

The broader contribution of this approach is mainly methodological. It offers a way to move from individual cases to a structured comparison of trajectories, read as movements through a field that also develops over time. Each measure specifies which relation between author and field it captures and the publication evidence on which the comparison rests. Historical interpretations can therefore be examined across authors against the same moving field baseline, with their evidential limits made explicit. The comparison can be extended to other fields where corpus coverage, metadata, and historically defensible field boundaries support it. For computational historiography, the relations between these partial traces are themselves an object of historical investigation. Their agreement and divergence locate where a general account of field change needs to be differentiated at the level of individuals.

\section*{Data Availability Statement}

The computed outputs of the analysis, the corpus identity (bibcode lists and per-record content checksums), the run configurations, the figure sources, and the rendering scripts are openly available as a reproduction package at Zenodo: \href{https://doi.org/10.5281/zenodo.22726750}{https://doi.org/10.5281/zenodo.22726750}. The analysis software is available as the \textit{Trajectories of Change} package, version 0.2.0, at Zenodo (\href{https://doi.org/10.5281/zenodo.22084879}{https://doi.org/10.5281/zenodo.22084879}) and on GitHub (\url{https://github.com/raphschlatt/Trajectories_of_Change}). The corpus-preparation and author-disambiguation pipelines are public at \url{https://github.com/raphschlatt/ads-bib} and \url{https://github.com/raphschlatt/ads-and}. The corpus records themselves were retrieved from the NASA Astrophysics Data System and are not redistributed. The reproduction package documents their exact identity and retrieval, and the frozen prepared corpus is available from the authors on reasonable request.

\bibliography{references}

\clearpage

%% ----------------------------------------------------------------------
%% Supplementary Material (content identical to the supplement of the
%% journal submission), typeset as part of this preprint so that the
%% main text and the supplement share one layout.
%% ----------------------------------------------------------------------
\setcounter{section}{0}
\setcounter{table}{0}
\setcounter{figure}{0}
\setcounter{equation}{0}
\renewcommand{\thesection}{S\arabic{section}}
\renewcommand{\thetable}{S\arabic{table}}
\renewcommand{\thefigure}{S\arabic{figure}}
\renewcommand{\theequation}{S\arabic{equation}}
%% Unique hyperref anchors: without these the reset counters collide with
%% the main-text anchors (section.1 etc.) and the PDF bookmarks jump wrong.
\renewcommand{\theHsection}{S\arabic{section}}
\renewcommand{\theHsubsection}{\theHsection.\arabic{subsection}}
\renewcommand{\theHtable}{S\arabic{table}}
\renewcommand{\theHfigure}{S\arabic{figure}}
\renewcommand{\theHequation}{S\arabic{equation}}

\section*{Supplementary Material}

This Supplementary Material is the complete formal specification of the analysis pipeline,
followed by the run parameters, the full Top-50 list, the embedding-density settings, the
counting-variant robustness, the reception-layer weighting, and the transfer-corpus
correlations. The specification is self-contained and therefore restates the compact
equations of the main text in full derivation form, in the same notation. All analyses were computed
with the \emph{Trajectories of Change} package (v0.2.0, DOI 10.5281/zenodo.22084879), whose implementation the equations
below follow.
Throughout, \(T\) is the target and \(F\) the field, \(s\) is a target slice and \(\tau\) a field slice. A feature \(x\) is a lemmatised term \(t\) for the vocabulary measures and an
unordered co-cited author pair \(\pi\) for Citation Identity, drawn from a measure-specific
feature set \(V\).

\section{Formal specification of the pipeline}
\label{sec:sm-spec}

\subsection{Data preparation}

Two tables enter the analysis, a publication table and a reference table keyed by ADS
\texttt{Bibcode}. Before any metric is computed, bibcodes are normalised, empty keys are
dropped, and duplicate keys are reduced to one row deterministically, keeping the row with
the longest reference list, then the most non-empty fields, then the first occurrence. No content is averaged. Within each publication, empty and duplicate reference identifiers are
removed, and a reference whose bibcode is absent from the reference table is dropped from
that list while the citing publication is kept, since an unresolved reference cannot supply
a cited author or text but the publication still contributes its own vocabulary and
position. Author identities are separated from display strings. The analytic field
\texttt{author\_uids} carries the stable identifiers produced by the purpose-built
disambiguation pipeline (\texttt{ads-and}, operating on corpus data prepared with
\texttt{ads-bib}, both public at \url{https://github.com/raphschlatt/ads-and} and
\url{https://github.com/raphschlatt/ads-bib}), is de-duplicated within a row, and is
stripped of placeholder markers (any uid containing \texttt{::n.author::},
\texttt{::unknown::}, ``unknown'', or ``no author'', and empty or whitespace uids), while
the raw \texttt{Author} display is retained. These steps remove non-analysable identity and
reference artefacts and correct no research content. For the reported corpus, 183{,}688
publication records and 445{,}843 reference records went in, and 183{,}680 and 445{,}832
came out. The preparation reduced 8 duplicate publication bibcodes and 11 duplicate
reference bibcodes, and it removed 202 unresolved reference identifiers (53 unique) from
reference lists while keeping the citing publications. It also removed 208 duplicate and
2{,}732 placeholder \texttt{author\_uids} from publication rows (483 and 684 from reference
rows), leaving 2{,}709 publication rows and 604 reference rows without a usable identifier.
Titles and abstracts are language-detected with fastText \texttt{lid.176} and, where
non-English, machine-translated to English (via \texttt{google/gemini-3-flash-preview} on
OpenRouter) before lemmatisation. The roster's \texttt{Lang} column reports the dominant
pre-translation title language, so the corpus is read in a common language while the original language
remains an auditable diagnostic.

\subsection{Slice distributions and Jelinek--Mercer smoothing}

The three distributional measures share one comparison. A target and the field are each
represented as a probability distribution over the shared feature set \(V\) within a time
slice. The estimate is formed at the level of the slice, not the individual document:
writing \(c_{T,s}(x)\) for the total count of feature \(x\) over the target's documents in
slice \(s\),
\begin{equation}
P_T(x \mid s) = \frac{c_{T,s}(x)}{\sum_{u \in V} c_{T,s}(u)} ,
\qquad
P_F(x \mid \tau) = \frac{c_{F,\tau}(x)}{\sum_{u \in V} c_{F,\tau}(u)} ,
\label{eq:mle}
\end{equation}
where the field counts are the corpus counts with the target removed,
\(c_{F,\tau}=c_{\mathrm{corpus},\tau}-c_{T,\tau}\). A single unseen feature gives a zero
probability and an infinite divergence, and absence from a short or weakly indexed slice is
not absence from the field. Each distribution is therefore interpolated with a background
distribution \(P_C\) and floored,
\begin{equation}
\widetilde{P}_T(x \mid s) \;\propto\; \max\!\Big[(1-\lambda)\,P_T(x \mid s) + \lambda\,P_C(x),\ \varepsilon\Big],
\label{eq:jm}
\end{equation}
renormalised to sum to one (blend, then floor, then renormalise), and likewise for the
field \(\widetilde{P}_F(x \mid \tau)\). The background \(P_C\) is the time-pooled field
distribution, that is the corpus counts minus the target counts summed over all slices,
itself floored and renormalised once. The interpolation weight is \(\lambda=0.05\) in the
main analysis and \(\lambda=0.5\) as a robustness axis, with floor \(\varepsilon=10^{-12}\).

\subsection{Divergence, synchronous and asynchronous comparison}

The distance between the smoothed target and field models is the Kullback--Leibler
divergence, computed in bits (base-2 logarithm),
\begin{equation}
D_{\mathrm{KL}}\big(\widetilde{P}_T(\cdot\mid s) \parallel \widetilde{P}_F(\cdot\mid \tau)\big)
= \sum_{x \in V} \widetilde{P}_T(x\mid s)\,\log_2 \frac{\widetilde{P}_T(x\mid s)}{\widetilde{P}_F(x\mid \tau)} .
\label{eq:kld}
\end{equation}
We write \(d_i^m(s,\tau)\) for this divergence for author \(i\) under measure \(m\). It is
used in two ways. The \emph{synchronous} value sets the field slice to the target slice,
\(d_i^m(s,s)\), and measures how far a target stands from the field of its own moment. The
\emph{asynchronous} comparison evaluates the same target slice against the field model of
every available slice \(\tau\), \(d_i^m(s,\tau)\), and records the closest one,
\begin{equation}
\tau_i^{m\ast}(s) = \arg\min_{\tau} d_i^m(s,\tau),
\qquad
\delta_i^m(s) = \tau_i^{m\ast}(s) - s ,
\label{eq:leadlag}
\end{equation}
the slice-level lead/lag \(\delta_i^m(s)\) in years. A negative value marks a target slice
that most resembles an earlier field state, a positive value a later one, and a value near
zero the contemporary field. These are similarity relations, not claims of influence,
priority, or anticipation. The field-slice search runs over every slice that clears the
support thresholds, so the differences in reach among measures are empirical.

\subsection{The divergence decomposition and its aggregates}

Because Eq.~\eqref{eq:kld} is a sum, it decomposes into signed per-feature contributions
\(D_x = \widetilde{P}_T(x\mid s)\log_2[\widetilde{P}_T(x\mid s)/\widetilde{P}_F(x\mid \tau)]\),
positive where the target over-uses \(x\) and negative where it under-uses it. Suppose a
term such as ``tetrad'' has smoothed probability \(0.05\) for the target and \(0.001\) for the field. Its contribution is \(0.05\cdot\log_2 50 \approx 0.28\). A field-wide term such
as ``Einstein'' (\(\approx0.03\) versus \(\approx0.02\)) contributes only
\(0.03\cdot\log_2 1.5 \approx 0.017\), so prominence is not distinctiveness. Three aggregates are reported per slice. The all-term divergence is the full sum,
\begin{equation}
\mathrm{KLD}_{\mathrm{all}} = \sum_{x \in V} D_x ,
\label{eq:kldall}
\end{equation}
which is the quantity used throughout the main text. With \(S\subseteq V\) the set of
document-stable features selected by the test of the next subsection,
\begin{equation}
\mathrm{KLD}_{\mathrm{sig}} = \sum_{x \in S} D_x ,
\qquad
\mathrm{KLD}_{\mathrm{sig,abs}} = \sum_{x \in S} |D_x| .
\label{eq:kldsig}
\end{equation}
Because contributions are signed, \(\mathrm{KLD}_{\mathrm{sig}}\) can cancel toward zero even
when stable differences exist (\(+0.05\) and \(-0.05\) give
\(\mathrm{KLD}_{\mathrm{sig}}\approx0\) but \(\mathrm{KLD}_{\mathrm{sig,abs}}\approx0.10\)),
so the signed sum is read as a direction and the absolute sum as the magnitude of stable
distinguishing features.

\subsection{Document-level Welch test and false-discovery control}

The candidate family in a slice pair is the fifty features with the largest \(|D_x|\). Fifty
keeps the family wide enough to cover the features that drive a slice pair's divergence while
keeping per-document testing meaningful under sparse support, and the all-term results do not
depend on this choice. For
each candidate the per-document relative frequency \(r_d(x)=c_d(x)/\ell_d\) is formed, where \(\ell_d\) is the document's in-vocabulary token count. Documents with no in-vocabulary tokens are dropped, and documents with tokens but without the feature contribute
\(r_d(x)=0\) and remain in the count. A Welch two-sample \(t\)-test (unequal variances, via
\texttt{scipy.stats.ttest\_ind\_from\_stats} with sample standard deviations,
\(\mathrm{ddof}=1\)) compares the \(N_T\) target and \(N_F\) field documents,
\begin{equation}
t = \frac{\bar{r}_T-\bar{r}_F}{\sqrt{\,s_T^2/N_T + s_F^2/N_F\,}},
\qquad
\nu \approx \frac{\big(s_T^2/N_T + s_F^2/N_F\big)^2}
{\dfrac{(s_T^2/N_T)^2}{N_T-1} + \dfrac{(s_F^2/N_F)^2}{N_F-1}} ,
\label{eq:welch}
\end{equation}
with \(N_T,N_F\) the full analysable-document counts per side. The two-sided \(p\)-values are
adjusted within a slice (synchronous) or a target slice (asynchronous) by the
Benjamini--Hochberg procedure, and a feature enters \(S\) when its adjusted value clears
\(q\le0.20\) in the main view or \(q\le0.10\) as a stricter view. A term used consistently
across documents has a small variance and a significant test. A term concentrated in one unusual document has a large variance and fails, so a single outlier work does not become a
stable marker. Without correction the screen would be swamped. At \(m=50\) tests and
\(\alpha=0.2\), about ten false positives per slice, and some \(450\) per author over the
\(45\) slices, would arise by chance.

\subsection{The four feature spaces}

Each measure instantiates the feature set \(V\) and the count \(c_{T,s}\) of Eq.~\eqref{eq:mle}. The smoothing, divergence, decomposition, and Welch test are then identical to the subsections above.

\paragraph{Own Vocabulary.} The features are the lemmatised terms \(t\) of the target's own
titles and abstracts, with \(c^{\mathrm{own}}_{T,s}(t)\) the term's count across the target's
publications in slice \(s\), so
\begin{equation}
P_T^{\mathrm{own}}(t \mid s)
= \frac{c^{\mathrm{own}}_{T,s}(t)}{\sum_{u \in V_{\mathrm{own}}} c^{\mathrm{own}}_{T,s}(u)} .
\label{eq:own}
\end{equation}

\paragraph{Referenced Vocabulary.} The features are the terms of the works the target cites,
under a two-level weighting. Within a cited work \(w\) a term carries its relative frequency \(\mathrm{tf}(t,w)/\ell_w\), and within a citing publication \(d\) the \(k_d\) usable cited works
share its mass equally. The referenced-vocabulary count is the accumulation of these weights
over the target's citing publications in slice \(s\),
\begin{equation}
c^{\mathrm{ref}}_{T,s}(t) = \sum_{d}\ \sum_{w \in d} \frac{1}{k_d}\cdot\frac{\mathrm{tf}(t,w)}{\ell_w} ,
\qquad
P_T^{\mathrm{ref}}(t \mid s)
= \frac{c^{\mathrm{ref}}_{T,s}(t)}{\sum_{u \in V_{\mathrm{ref}}} c^{\mathrm{ref}}_{T,s}(u)} ,
\label{eq:ref}
\end{equation}
so each citing publication contributes total mass \(1\) and neither a long bibliography nor a
long cited abstract dominates. The slice axis is the citing publication's year. The inclusive
policy keeps the target's own cited works. The external-only policy removes them before counting, reducing \(k_d\).

\paragraph{Citation Identity.} The features are unordered pairs \(\pi\) of co-cited authors,
taking the first listed author (disambiguated \texttt{author\_uid} where available) of each
cited work. Within a citing document the surviving distinct pairs are counted under one of
three models, which set the per-document weight of a pair. Document-fractional counting, the canonical model, gives each of the \(n_d\) distinct post-filter pairs weight \(1/n_d\), so the document contributes total mass \(1\). Binary counting gives each distinct pair weight \(1\), and multiplicity counting gives a pair the number of distinct cited-work combinations that produce it. Same-author
pairs are removed before counting, and under the canonical target exclusion
(\texttt{all\_docs}) a pair is dropped from every document when either of its two source
references is by the target, so the field baseline does not contain the target's own incoming
reception. Writing \(c^{\mathrm{cit}}_{T,s}(\pi)\) for the accumulated pair weight across the
target's citing publications in slice \(s\), the profile and divergence are formed exactly as
for the vocabulary measures,
\begin{equation}
P_T^{\mathrm{cit}}(\pi \mid s)
= \frac{c^{\mathrm{cit}}_{T,s}(\pi)}{\sum_{\pi' \in V_{\mathrm{pair}}} c^{\mathrm{cit}}_{T,s}(\pi')} ,
\label{eq:cit}
\end{equation}
and Eq.~\eqref{eq:kld} is applied to \(\widetilde{P}_T^{\mathrm{cit}}\) and
\(\widetilde{P}_F^{\mathrm{cit}}\) over the pair space \(V_{\mathrm{pair}}\). The dropped
self-loop, target-excluded, and unresolved pair mass is reported per author and slice as a
diagnostic, and a global pair-support threshold guards against pairs carried by too few
documents.

\paragraph{Embedding Density.} The features are not discrete. Each publication is embedded
and projected to low-dimensional coordinates (Table~\ref{tab:ede}), standardised once over
the whole corpus by a global \(z\)-score (population standard deviation, \(\mathrm{ddof}=0\)).
The field density at a coordinate \(z\) in slice \(\tau\) is a Gaussian kernel-density
estimate fitted on the field publications \(\mathcal{D}_{F,\tau}\) (the corpus of that slice
with the target removed),
\begin{equation}
\rho_F(z \mid \tau) = \frac{1}{|\mathcal{D}_{F,\tau}|}\sum_{q\in\mathcal{D}_{F,\tau}} K_h(z-z_q),
\qquad
h = n^{-1/(k+4)} ,
\label{eq:kde}
\end{equation}
where \(K_h\) is the Gaussian kernel, \(k\) the coordinate dimension, and the Scott bandwidth
\(h\) uses the global corpus size \(n\) computed once, so the same \(h\) is reused for every
slice. The slice score is the median over the target's publications of the negative
\emph{natural} logarithm of the density,
\begin{equation}
\mathrm{EDE}(s\mid\tau) = \operatorname{median}_{p}\big[-\ln \rho_F(z_p \mid \tau)\big] ,
\label{eq:ede}
\end{equation}
so a lower score marks a denser neighbourhood. It enters the synchronous and asynchronous
machinery of Eqs.~\eqref{eq:kld}--\eqref{eq:leadlag} as \(d_i^{\mathrm{EDE}}(s,\tau)\), with
no Welch analogue since it has no discrete features.

\subsection{Author-level aggregation}

For the synchronous series \(x_i^m(s)=d_i^m(s,s)\), the reported author-level summaries are
the median level, the ordinary-least-squares slope, and the lead/lag,
\begin{equation}
L_i^m = \operatorname{median}_s x_i^m(s),
\qquad
\beta_i^m = \operatorname{slope}_s x_i^m(s),
\qquad
\Delta_i^m = \operatorname{mean}_s\, \delta_i^m(s) ,
\label{eq:agg}
\end{equation}
with \(\delta_i^m(s)\) from Eq.~\eqref{eq:leadlag} and the slope undefined for fewer than two
active slices. The lead/lag and its matching level are unweighted means over the active
slices, not medians. Cross-author consistency is read from Spearman rank correlations of
\(L\), \(\beta\), and \(\Delta\) separately, under the bands stated in the main text.

\subsection{Design choices and their historical reading}

The measures share a handful of design choices whose historical reading is not obvious from
their definitions. Table~\ref{tab:failure-modes} collects the most important ones together
with the problem each avoids.

\begin{table}[ht!]
\caption{Design choices shared by the measures of Section~\ref{sec:sm-spec}, the problem each avoids, and its historical reading.}
\label{tab:failure-modes}
\begin{center}
\footnotesize
\begin{tabular}{>{\raggedright\arraybackslash}p{0.19\textwidth}>{\raggedright\arraybackslash}p{0.24\textwidth}>{\raggedright\arraybackslash}p{0.23\textwidth}>{\raggedright\arraybackslash}p{0.22\textwidth}}
\hline
Choice & What it does & Problem avoided & Historical reading \\
\hline
Jelinek--Mercer smoothing & Gives a term absent from a thin slice a small background mass. & Infinite or unstable divergence from zero probabilities. & Absence from a weakly indexed slice is not read as absence from the field. \\
External-only Referenced Vocabulary & Drops the target's own cited works from the cited-language profile. & Cited vocabulary silently recycling the author's own wording. & Separates the cited horizon from self-citation. \\
Document-fractional weighting & Gives each citing paper the same total mass. & Long bibliographies dominating the profile by length. & A review article does not outweigh a short paper by citing more. \\
Same-author loop removal & Drops pairs that join an author with themselves. & One repeated authority posing as relational structure. & Citation Identity tracks pairings, not single-author anchoring. \\
Target exclusion & Removes the target's own works from the field baseline. & Outgoing reference practice blurring with incoming reception. & Keeps Citation Identity distinct from Citation Image. \\
First-author scope & Forms pairs from the first author of each cited work. & All-author expansion generating many pair events per paper. & Cuts noise but can hide collaborative citation structure (Section~\ref{sec:sm-counting}). \\
Welch/FDR feature layer & Keeps only document-stable distinguishing terms or pairs. & A single unusual document driving the interpretation. & A diagnostic layer, not a second classification of authors. \\
Negative-log density (EDE) & Reports \(-\log\) field density, so the value is distance-like. & Confusing raw density with a distance-like score. & Lower means denser; EDE signs read opposite to raw density. \\
\hline
\end{tabular}
\end{center}
\end{table}

\section{Reproducibility parameters}

Table~\ref{tab:repro} lists the parameters of the reported Top-50 run. The exact ADS query
is the set-based construction of the earlier study. It is included in the public code package as a tracked file (\texttt{docs/ads\_query.md}), together with the release commit, and is not restated here.

\begin{table}[ht!]
\centering
\caption{Run parameters for the canonical ADS-GRG Top-50 analysis. Values are read from the released run configuration and the primary metric matrix and are identical across all fifty authors. Items marked ``with code release'' are deterministic inputs shipped with the software package, not free parameters.}
\label{tab:repro}
\small
\begin{tabular}{@{}p{0.40\textwidth}p{0.52\textwidth}@{}}
\toprule
Parameter & Value \\
\midrule
\multicolumn{2}{@{}l}{\textit{Corpus and prepared data}}\\
Source & NASA/ADS, GRG publications, 1911--2000 (183{,}680 records) \\
ADS query strategy & set-based construction of the earlier study (with code release) \\
Prepared dataset & \texttt{grg\_ads\_toponymy\_20260430\_explore\_ref\_tokens} \\
\midrule
\multicolumn{2}{@{}l}{\textit{Time slices}}\\
Window size & 2 years, non-overlapping \\
Field slices & 45 over the corpus span \\
\midrule
\multicolumn{2}{@{}l}{\textit{Smoothing}}\\
Jelinek--Mercer weight $\lambda$ & 0.05 (main); 0.5 (robustness) \\
Probability floor $\varepsilon$ & $10^{-12}$ \\
\midrule
\multicolumn{2}{@{}l}{\textit{Document-level (Welch/FDR) layer}}\\
Candidate set & top 50 features by $|D_x|$ per slice pair \\
Test & Welch two-sample $t$ (unequal variance, $\mathrm{ddof}=1$) \\
Multiple testing & Benjamini--Hochberg (\texttt{fdr\_bh}) \\
False-discovery rate $q$ & 0.20 (main); 0.10 (sensitivity) \\
Scope & \texttt{slice} (synchronous); \texttt{target\_slice} (asynchronous) \\
Near-field band & author-mean lead/lag within $\pm 1$ year (half a slice step) \\
\midrule
\multicolumn{2}{@{}l}{\textit{Citation Identity}}\\
Feature & unordered cited-author pairs (\texttt{cocit\_mode = authors}) \\
Counting & document-fractional (mass $1.0$ per citing document) \\
Author scope & first author of each cited work \\
Target exclusion & all documents (\texttt{all\_docs}) \\
Self-loops & removed \\
\midrule
\multicolumn{2}{@{}l}{\textit{Referenced Vocabulary}}\\
Policy & inclusive (main); external-only (self-citation check) \\
Weighting & two-level (Eq.~\eqref{eq:ref}) \\
Time axis & year of the citing publication \\
Vocabulary cap / floor & $50{,}000$ terms; minimum global document frequency $2$ \\
\midrule
\multicolumn{2}{@{}l}{\textit{Embedding Density}}\\
Settings & see Table~\ref{tab:ede} \\
\bottomrule
\end{tabular}
\end{table}

\section{The Top-50 list}

Table~\ref{tab:roster} is the corpus-visible portfolio on which the correlations,
orientation counts, and trajectory families rest. It is a coverage portfolio, not an
excellence canon. Reference support per publication ranges from below two resolved
references (Treder) to above thirty (Olive, Nanopoulos), and is especially thin for the
instrument and Soviet cases (Weber, Pizzella, Zel'dovich, Novikov). One author, Fritts, enters through the keyword boundary alone, with atmospheric gravity-wave research in the \emph{Journal of the Atmospheric Sciences}. Aggregate findings shift by at most one
author when such a boundary case is removed.

{\footnotesize
\begin{longtable}{@{}lrrrrrrlrl@{}}
\caption{The ADS-GRG Top-50 roster, ordered by corpus publications. Pubs: matched
publications. Years: first and last active year. Sl.: active two-year slices. Pubs(r):
publications carrying resolved references. Refs: total resolved references. R/p: resolved
references per publication. Lang: dominant title language. Recep.: field-internal future
no-self reception events (Section~\ref{sec:sm-citimage}). Flags: I marks an author-identity mixture inside the
matched identifier, T off-domain topic contamination, B a keyword-boundary case, and n no
usable Welch/FDR feature support at \(q \le 0.20\) in at least one layer (Hoyle: Own
Vocabulary; Barrow: Citation Identity; Weber: external-only Referenced Vocabulary).}
\label{tab:roster}\\
\toprule
Author & Pubs & Years & Sl. & Pubs(r) & Refs & R/p & Lang & Recep. & Flags \\
\midrule
\endfirsthead
\multicolumn{10}{@{}l}{\textit{Table~\ref{tab:roster} continued}}\\
\toprule
Author & Pubs & Years & Sl. & Pubs(r) & Refs & R/p & Lang & Recep. & Flags \\
\midrule
\endhead
\midrule
\multicolumn{10}{r@{}}{\textit{continued on next page}}\\
\endfoot
\bottomrule
\endlastfoot
Treder, H.\,J. & 350 & 1957--2000 & 22 & 155 & 639 & 1.8 & de & 276 & -- \\
Silk, Joseph & 330 & 1966--2000 & 18 & 221 & 7081 & 21.5 & en & 4739 & -- \\
Schramm, David N. & 320 & 1967--1999 & 15 & 140 & 4649 & 14.5 & en & 3523 & -- \\
Narlikar, J.\,V. & 252 & 1961--2000 & 20 & 129 & 2308 & 9.2 & en & 915 & -- \\
Turner, Michael S. & 252 & 1977--2000 & 12 & 146 & 5527 & 21.9 & en & 4188 & -- \\
Shapiro, I.\,I. & 239 & 1964--2000 & 19 & 143 & 2521 & 10.5 & en & 2124 & I \\
Rees, M.\,J. & 233 & 1966--2000 & 18 & 139 & 4323 & 18.6 & en & 5046 & -- \\
Nakamura, T. & 229 & 1957--2000 & 15 & 161 & 3261 & 14.2 & en & 1067 & -- \\
Odintsov, S.\,D. & 227 & 1982--2000 & 10 & 185 & 4361 & 19.2 & en & 465 & -- \\
Anderson, J.\,D. & 219 & 1951--2000 & 25 & 109 & 1799 & 8.2 & en & 861 & I \\
Ellis, G.\,F.\,R. & 199 & 1964--2000 & 19 & 129 & 3307 & 16.6 & en & 2299 & -- \\
Barrow, John D. & 199 & 1976--2000 & 13 & 143 & 3763 & 18.9 & en & 2474 & n \\
Novikov, I.\,D. & 196 & 1961--2000 & 20 & 61 & 1324 & 6.8 & en & 1619 & -- \\
Shapiro, Stuart L. & 194 & 1972--2000 & 15 & 156 & 3870 & 19.9 & en & 2301 & -- \\
Ostriker, Jeremiah P. & 193 & 1964--2000 & 19 & 144 & 5221 & 27.1 & en & 5728 & -- \\
Deser, S. & 188 & 1953--2000 & 23 & 134 & 1381 & 7.3 & en & 3432 & -- \\
Nanopoulos, D.\,V. & 185 & 1973--2000 & 13 & 147 & 5725 & 30.9 & en & 2223 & -- \\
Hawking, S.\,W. & 184 & 1965--2000 & 18 & 95 & 1300 & 7.1 & en & 8354 & -- \\
Damour, T. & 182 & 1974--2000 & 13 & 92 & 2270 & 12.5 & en & 1792 & -- \\
Schutz, B.\,F. & 182 & 1970--2000 & 16 & 84 & 1281 & 7.0 & en & 1296 & -- \\
Melnikov, V.\,N. & 182 & 1972--2000 & 15 & 94 & 1563 & 8.6 & en & 235 & -- \\
Ellis, John & 178 & 1966--2000 & 13 & 140 & 5035 & 28.3 & en & 1918 & -- \\
Zel'dovich, Ya.\,B. & 175 & 1963--1998 & 17 & 56 & 1621 & 9.3 & en & 3611 & -- \\
Thorne, Kip S. & 174 & 1965--2000 & 18 & 82 & 1732 & 10.0 & en & 6772 & -- \\
Smith, D.\,E. & 169 & 1956--2000 & 21 & 86 & 1462 & 8.7 & en & 442 & -- \\
Olive, Keith A. & 169 & 1978--2000 & 12 & 123 & 5389 & 31.9 & en & 2912 & -- \\
Hoyle, F. & 166 & 1939--2000 & 26 & 82 & 1185 & 7.1 & en & 1090 & n \\
Mann, R.\,B. & 163 & 1980--2000 & 11 & 131 & 2952 & 18.1 & en & 1183 & -- \\
Peebles, P.\,J.\,E. & 163 & 1962--2000 & 20 & 117 & 3152 & 19.3 & en & 8578 & -- \\
Ruffini, R. & 163 & 1968--2000 & 17 & 92 & 1381 & 8.5 & en & 684 & -- \\
Fang, LiZhi & 160 & 1974--2000 & 14 & 76 & 1869 & 11.7 & en & 448 & -- \\
Maeda, KeiIchi & 155 & 1975--2000 & 13 & 121 & 3166 & 20.4 & en & 1669 & -- \\
Wesson, P.\,S. & 155 & 1973--2000 & 14 & 112 & 2669 & 17.2 & en & 801 & -- \\
Pizzella, G. & 154 & 1974--2000 & 14 & 64 & 448 & 2.9 & en & 379 & -- \\
Smith, R.\,C. & 150 & 1957--2000 & 19 & 91 & 2224 & 14.8 & en & 1063 & -- \\
Turner, E.\,L. & 149 & 1975--2000 & 13 & 98 & 2699 & 18.1 & en & 2858 & -- \\
Ellis, R.\,S. & 147 & 1977--2000 & 12 & 91 & 3106 & 21.1 & en & 2559 & I \\
White, S.\,D.\,M. & 146 & 1976--2000 & 13 & 119 & 4040 & 27.7 & en & 6429 & -- \\
Linde, A.\,D. & 146 & 1974--2000 & 13 & 89 & 3324 & 22.8 & en & 3974 & -- \\
Matzner, Richard A. & 145 & 1967--2000 & 17 & 101 & 1857 & 12.8 & en & 1095 & -- \\
Kolb, Edward W. & 144 & 1977--2000 & 12 & 67 & 1908 & 13.3 & en & 1804 & -- \\
Sasaki, Misao & 144 & 1972--2000 & 14 & 99 & 2049 & 14.2 & en & 1420 & -- \\
Davis, Marc & 142 & 1964--2000 & 17 & 103 & 3259 & 23.0 & en & 5996 & I \\
Lee, C.\,C. & 140 & 1971--2000 & 15 & 79 & 1426 & 10.2 & en & 200 & -- \\
Schneider, P. & 140 & 1983--2000 & 9 & 89 & 2415 & 17.3 & en & 1935 & -- \\
Piran, Tsvi & 139 & 1975--2000 & 13 & 97 & 2407 & 17.3 & en & 1678 & -- \\
Wheeler, J.\,C. & 139 & 1955--2000 & 23 & 62 & 1262 & 9.1 & en & 4602 & I \\
Weber, J. & 138 & 1957--1999 & 22 & 44 & 302 & 2.2 & en & 921 & T, n \\
Gibbons, G.\,W. & 137 & 1971--2000 & 15 & 104 & 2026 & 14.8 & en & 3619 & -- \\
Fritts, David C. & 136 & 1975--2000 & 13 & 85 & 2390 & 17.6 & en & 1722 & B \\
\end{longtable}
}

\section{Embedding Density settings}

The document and reference texts are embedded through OpenRouter and reduced to the analysis
coordinates by PaCMAP. The Qwen layers embed both document and reference texts with the Qwen3
8B embedding model, and the Gemini layer embeds documents with the Gemini embedding model
while keeping the Qwen3 reference embeddings. Table~\ref{tab:ede} gives the exact model identifiers and per-layer settings. The kernel-density estimate is standardised
once over the whole corpus and fitted on the field publications of each slice, with the
target publications split out of the field they are scored against. Because each slice's
density integrates to one, the asynchronous comparison across slices compares the share of a field slice's publications near the target's publications. Small, compact early slices reach higher peak densities than large, spread-out later ones, which the main text notes among its
limits. A within-slice rank of density is reserved for a follow-up analysis.

\begin{table}[ht!]
\centering
\caption{Embedding Density layers. The Qwen 2D layer is primary. Qwen 5D and Gemini 2D are robustness checks. Densities in different coordinate spaces are not on the same numerical
scale and are compared only within a layer.}
\label{tab:ede}
\footnotesize
\begin{tabular}{@{}p{0.235\textwidth}p{0.225\textwidth}p{0.205\textwidth}p{0.205\textwidth}@{}}
\toprule
 & Qwen 2D (primary) & Qwen 5D (check) & Gemini 2D (check) \\
\midrule
Document embedding & \texttt{qwen/qwen3-}\newline\texttt{embedding-8b} & \texttt{qwen/qwen3-}\newline\texttt{embedding-8b} & \texttt{google/gemini-}\newline\texttt{embedding-001} \\
Reference embedding & \texttt{qwen/qwen3-}\newline\texttt{embedding-8b} & \texttt{qwen/qwen3-}\newline\texttt{embedding-8b} & \texttt{qwen/qwen3-}\newline\texttt{embedding-8b} \\
Provider & OpenRouter & OpenRouter & OpenRouter \\
Reduction & PaCMAP & PaCMAP & PaCMAP \\
Reduction neighbours & 80 & 60 & 80 \\
Reduction metric / seed & angular / 42 & angular / 42 & angular / 42 \\
Coordinates & \texttt{embedding\_2d\_\{x,y\}} & \texttt{embedding\_5d\_0..4} & \texttt{embedding\_2d\_\{x,y\}} \\
Dimension $k$ & 2 & 5 & 2 \\
Standardisation & global $z$, $\mathrm{ddof}=0$ & global $z$, $\mathrm{ddof}=0$ & global $z$, $\mathrm{ddof}=0$ \\
Kernel & Gaussian & Gaussian & Gaussian \\
Bandwidth & Scott, global ($h\approx0.133$) & Scott, global & Scott, global \\
Target policy & split out of field & split out of field & split out of field \\
\bottomrule
\end{tabular}
\end{table}

\section{Counting-variant robustness}
\label{sec:sm-counting}

Table~\ref{tab:robust} gives the Spearman correlations behind the robustness boundaries the
main text reports. Rank structure is largely preserved across the variants. The author-scope, counting, and target-exclusion sensitivities tabulated here are the ones the main text
summarises.

\begin{table}[ht!]
\centering
\caption{Citation Identity robustness against the canonical policy (document-fractional,
first-author, full target exclusion). Spearman $\rho$ across the fifty authors. Orientation
changes count authors whose slope or lead/lag sign flips relative to canonical. The target exclusion and self-loop rows are reported as slope sign flips in the source artifact.}
\label{tab:robust}
\small
\begin{tabular}{@{}lrrr@{}}
\toprule
Variant (vs canonical) & Slope $\rho$ & Lead/lag $\rho$ & Orientation changes \\
\midrule
\multicolumn{4}{@{}l}{\textit{Author scope}}\\
first-author $\rightarrow$ all-author & 0.712 & 0.682 & 16 slope; 24 lead/lag \\
\midrule
\multicolumn{4}{@{}l}{\textit{Counting model}}\\
document-fractional $\rightarrow$ binary & 0.673 & 0.669 & 14 slope; 22 lead/lag \\
document-fractional $\rightarrow$ multiplicity & 0.632 & 0.724 & 15 slope; 17 lead/lag \\
\midrule
\multicolumn{4}{@{}l}{\textit{Target exclusion and self-loops}}\\
all-docs $\rightarrow$ target-docs-only & 0.994 & --- & 0 slope sign flips \\
all-docs $\rightarrow$ no exclusion & 0.638 & --- & 6 slope sign flips \\
remove $\rightarrow$ retain self-loops & 0.967 & --- & 1 slope sign flip \\
\bottomrule
\end{tabular}
\end{table}

\section{Citation Image weighting}
\label{sec:sm-citimage}

Citation Image is reception, kept apart from the outgoing Citation Identity. With the target
author's publications \(\mathcal{P}\) and the observation endpoint \(Y=2000\),
\emph{field-internal future no-self reception} is the count of reference events in which a
later field publication \(c\) cites a publication \(p\in\mathcal{P}\),
\begin{equation}
N = \#\big\{(c,p): p\in\mathrm{refs}(c),\ \mathrm{year}(c)>\mathrm{year}(p),\
\mathrm{year}(c)\le Y,\ \mathrm{target}\notin\mathrm{authors}(c)\big\},
\label{eq:reception}
\end{equation}
with duplicate references to the same \(p\) within one citing document collapsed.
\emph{Coauthor-fractional Citation Count} weights each event by the inverse number of authors
of the cited publication,
\begin{equation}
N_{\mathrm{frac}} = \sum_{(c,p)} \frac{1}{\max\big(1,\ |\mathrm{authors}(p)|\big)} .
\label{eq:fractional}
\end{equation}
\emph{Exposure-normalised reception} divides by the available publication-years through the
endpoint,
\begin{equation}
N_{\mathrm{exp}} = \frac{N}{\sum_{p\in\mathcal{P}} \max\big(0,\ Y-\mathrm{year}(p)\big)} .
\label{eq:exposure}
\end{equation}
\emph{Received co-citation contexts} are formed per citing document from the other works it
cites alongside \(p\), excluding \(p\) and works authored by the target. Each context work receives weight \(1/|\text{context works}|\) and each context author \(1/|\text{context authors}|\) in that document, summed per target author and context. The no-target-self policy drops citing documents whose authors include the target.

\section{Transfer corpora}

The two control corpora, an INSPIRE-HEP holography corpus and a Semantic Scholar NLP/ACL corpus, are built with the same pipeline as the GRG analysis and reduced to a fifty-author control slice each. They test whether the measures can be constructed elsewhere and how their couplings behave. They are not evidence for a general law. Table~\ref{tab:control}
gives the cross-author Spearman correlations of the slope summaries.

\begin{table}[ht!]
\centering
\caption{Cross-author Spearman correlations of the slope summaries (\(\rho_{sl}\), as in Table~\ref{tab:slope-correlations} of the main text) among the measures in the two control corpora,
for the inclusive and external-only reference policies. Own: Own Vocabulary; Ref:
Referenced Vocabulary; CitId: Citation Identity; Den: Embedding Density. $n=50$ except the
ACL Citation-Identity pairs ($n=46$, authors without usable co-citation support excluded).}
\label{tab:control}
\small
\begin{tabular}{@{}lrrrr@{}}
\toprule
 & \multicolumn{2}{c}{INSPIRE holography} & \multicolumn{2}{c}{Semantic Scholar ACL} \\
\cmidrule(lr){2-3}\cmidrule(lr){4-5}
Measure pair & inclusive & external & inclusive & external \\
\midrule
Own--Ref     & 0.804 & 0.821 & 0.354 & 0.314 \\
Own--CitId   & 0.539 & 0.539 & 0.095 & 0.095 \\
CitId--Ref   & 0.537 & 0.510 & 0.041 & 0.056 \\
Own--Den     & 0.095 & 0.095 & 0.002 & 0.002 \\
CitId--Den   & $-0.100$ & $-0.100$ & 0.165 & 0.165 \\
Ref--Den     & 0.104 & 0.075 & 0.109 & 0.112 \\
\bottomrule
\end{tabular}
\end{table}

The Own--Ref coupling is strong in INSPIRE ($\rho\approx0.80$) and weak in ACL
($\rho\approx0.35$), and the reference- and density-based couplings fall to near-independence in ACL (Table~\ref{tab:control}). Under the inclusive policy both control corpora are
predominantly near-synchronous in lead/lag (INSPIRE 6 later, 27 near, 17 earlier of fifty
authors; ACL 3, 45, 2). The historical reading is given in the main text.

\section{Synchronous trajectory overview and case-author dashboards}

\begin{figure}[p]
\begin{center}
\includegraphics[width=\textwidth,height=0.90\textheight,keepaspectratio]{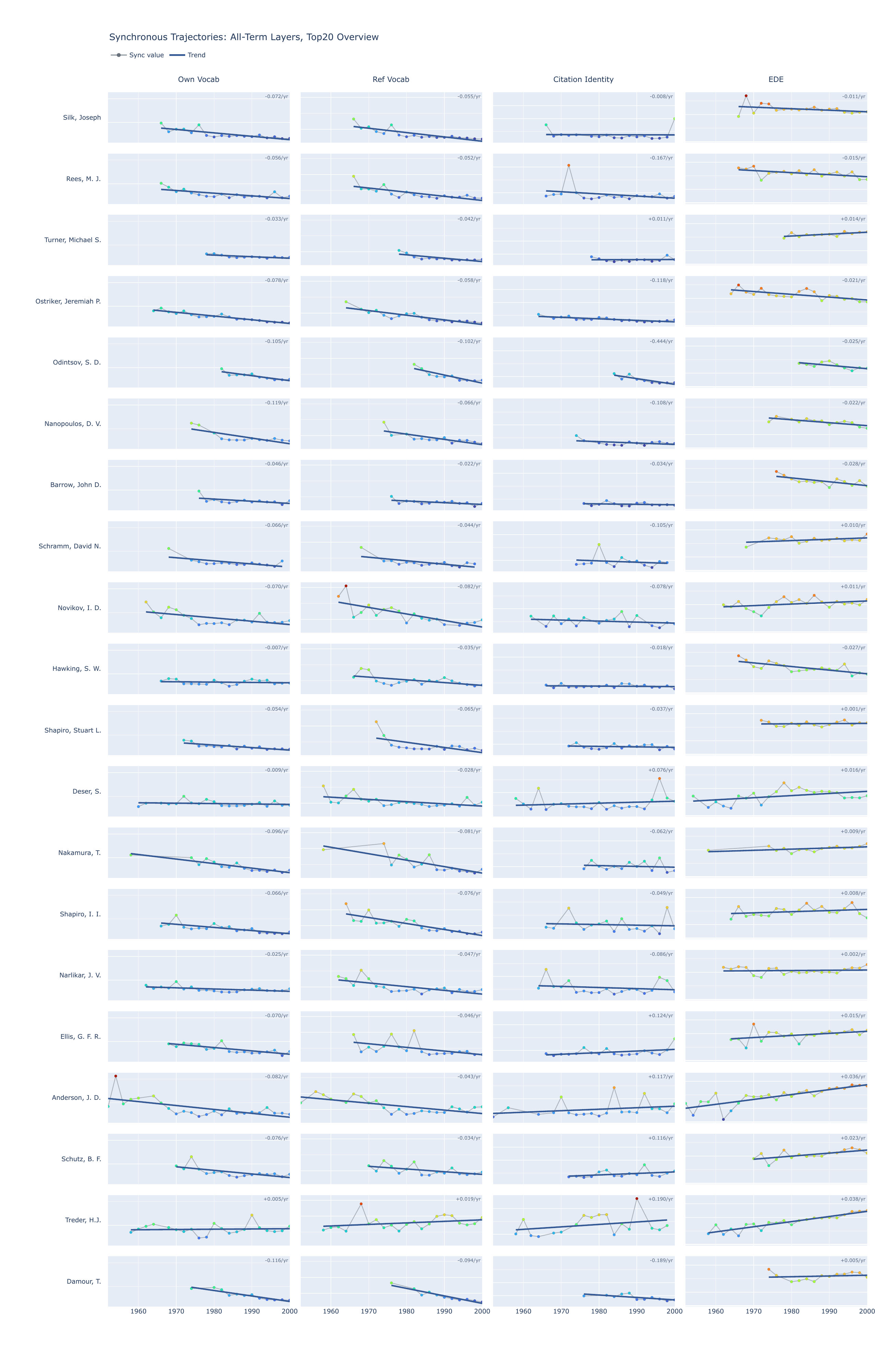}
\end{center}
\caption{Synchronous trajectories for the twenty most strongly covered of the Top-50
authors. Own Vocabulary, Referenced Vocabulary, and Citation Identity plot KLD from the contemporary field, and EDE plots negative-log field density. Sign conventions follow the lead/lag readings of the main text.}
\label{fig:sm-sync-overview}
\end{figure}

Figure~\ref{fig:sm-sync-overview} shows the synchronous trajectories for the twenty most
strongly covered authors, slice by slice, behind the orientation counts of the main text.
Own Vocabulary, Referenced Vocabulary, and Citation Identity plot KLD from the contemporary field, and EDE plots negative-log field density at the target publications' embedding coordinates. Declining KLD means that the author's distribution moves closer to the field
of the same moment, while lower EDE values indicate denser semantic neighbourhoods. These
synchronous trends are separate from the asynchronous lead/lag relations discussed in the
main text.

The per-author asynchronous dashboards for all fifty authors, in the format of the Silk and Treder
dashboards of the main text (referenced-vocabulary and density comparisons of each target slice
against earlier and later field slices, with the closest field slice marked), are part of the
reproduction package (see the Data Availability Statement of the main text).

\end{document}